\documentclass[10pt,twocolumn,letterpaper]{article}

\usepackage[pagenumbers]{cvpr} 

\usepackage{graphicx}
\usepackage{amsmath}
\usepackage{amssymb}
\usepackage{booktabs}
\usepackage[table,xcdraw,dvipsnames]{xcolor}

\usepackage[pagebackref,breaklinks,colorlinks]{hyperref}

\usepackage[capitalize]{cleveref}
\crefname{section}{Sec.}{Secs.}
\Crefname{section}{Section}{Sections}
\Crefname{table}{Table}{Tables}
\crefname{table}{Tab.}{Tabs.}

\def\cvprPaperID{*****} 
\def\confName{CVPR}
\def\confYear{2022}

\usepackage{graphicx}
\usepackage{tcolorbox}
\newcommand{\VarSty}[1]{\textnormal{\ttfamily\color{blue!90!black}#1}\unskip}

\usepackage{booktabs}
\usepackage{multirow}
\usepackage{graphicx}
\usepackage[normalem]{ulem}
\useunder{\uline}{\ul}{}

\usepackage{amsthm}
\usepackage{amsmath}
\usepackage{amsfonts}
\usepackage{xcolor}
\usepackage{mdframed}

\newtheoremstyle{fancydefinition}
  {\topsep}
  {\topsep}
  {\normalfont}
  {0pt}
  {\bfseries}
  {.}
  {0.5em}
  {}

\theoremstyle{fancydefinition}
\newtheorem{definition}{Definition}[section]
\mdfdefinestyle{coloredbox}{
  linecolor=blue!30!black,
  backgroundcolor=blue!1!white,
  roundcorner=5pt,
  innertopmargin=1pt,
  innerbottommargin=3pt,
  innerleftmargin=5pt,
  innerrightmargin=5pt,
  shadow=true,
  shadowcolor=black!10!white,
  shadowsize=2pt,
}

\newcommand{\tb}[3]{\setlength{\tabcolsep}{#2mm}\begin{tabular}{#1}#3\end{tabular}}

\begin{document}

\title{Visual Instruction Tuning with Polite Flamingo}

\author{
\tb{@{}cccc@{}}{5}{
Delong Chen$^{1,2}$ & 
Jianfeng Liu$^{1}$ & 
Wenliang Dai$^{2}$ &
Baoyuan Wang$^{1}$\\
}\\
\tb{cc}{5}{
$^{1}$Xiaobing.AI&  
$^{2}$Centre for Artificial Intelligence Research (CAiRE), HKUST
}}
\maketitle

\begin{abstract}
Recent research has demonstrated that the multi-task fine-tuning of multi-modal Large Language Models (LLMs) using an assortment of annotated downstream vision-language datasets significantly enhances their performance. Yet, during this process, a side effect, which we termed as the ``multi-modal alignment tax", surfaces. This side effect negatively impacts the model's ability to format responses appropriately - for instance, its ``politeness" - due to the overly succinct and unformatted nature of raw annotations, resulting in reduced human preference. In this paper, we introduce Polite Flamingo, a multi-modal response rewriter that transforms raw annotations into a more appealing, ``polite" format. Polite Flamingo is trained to reconstruct high-quality responses from their automatically distorted counterparts and is subsequently applied to a vast array of vision-language datasets for response rewriting. After rigorous filtering, we generate the PF-1M dataset and further validate its value by fine-tuning a multi-modal LLM with it. Combined with novel methodologies including U-shaped multi-stage tuning and multi-turn augmentation, the resulting model, Clever Flamingo, demonstrates its advantages in both multi-modal understanding and response politeness according to automated and human evaluations.\footnote{\scriptsize \url{https://github.com/ChenDelong1999/polite-flamingo}}
\end{abstract}

\section{Introduction}

\begin{figure}
    \centering
    \includegraphics[width=\linewidth]{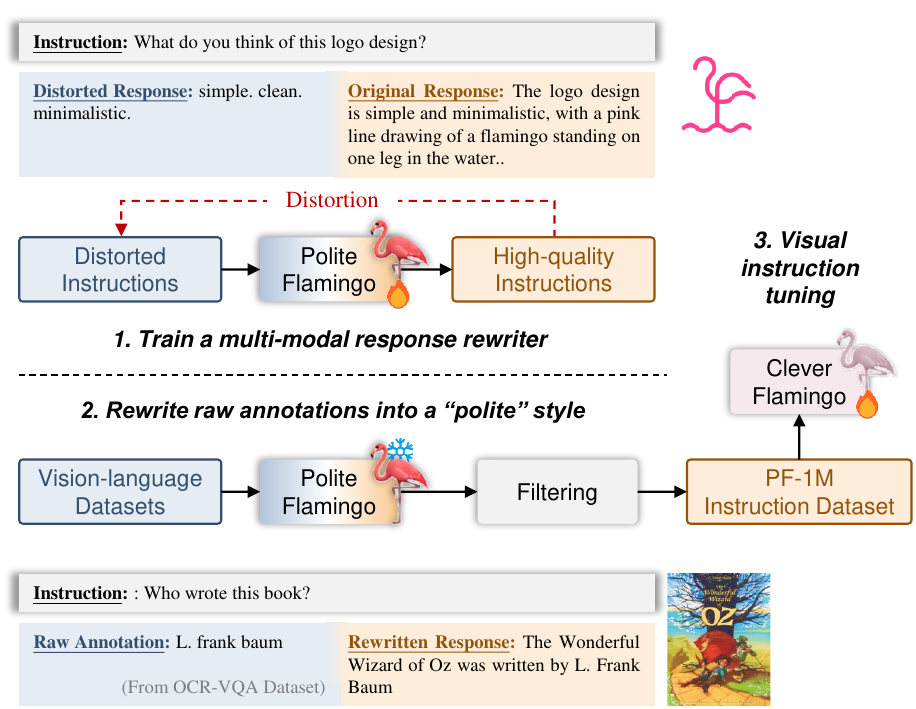}
    \caption{\textbf{Overview of our approach}. We first train a multi-modal response rewriter on high-quality instruction datasets, resulting in a ``Polite Flamingo'' capable of translating raw annotations in vision-language datasets into a ``polite'' style. After careful filtering, we use the rewritten data PF-1M for visual instruction tuning, and obtained a strong multi-modal LLM called Clever Flamingo.}
    \label{fig:overview}
\end{figure}

General-purpose AI systems have attracted a significant amount of interest due to their broad range of applications (\textit{e.g.}, smart assistants). They are expected to be capable of accurately perceiving the visual world, comprehending diverse human requests, and providing helpful yet natural responses.
Prior works towards this goal (\textit{e.g,} OFA~\cite{Wang2022OFA}, Unified-IO~\cite{Lu2022Unified}, Uni-Perceiver~\cite{Zhu2022Uni}) have focused on training multi-modal transformers via multi-task learning, but they lack the generalization ability to unseen tasks or instructions, and they are not capable of offering user-friendly natural responses. Recently, instruction tuning~\cite{Ouyang2022Training} empowers Large Language Models (LLMs)~\cite{Zhao2023Survey} strong instruction-following and response formatting abilities, making it more convenient and efficient to access its encoded knowledge and complex reasoning ability. Many researchers attempted to connect visual representations with LLMs to transfer such powerful capability to vision-language tasks. Massive image-text data collected from the Internet can be used to train the visual representation (\textit{e.g.,} CLIP~\cite{Radford2021Learning}) and the connector (\textit{e.g.,} Flamingo~\cite{Alayrac2022Flamingo}, Kosmos-1~\cite{Huang2023Language}, LLaVA~\cite{Liu2023Visual}, MiniGPT-4~\cite{Zhu2023MiniGPT}), but such supervision is usually noisy and could not cover much fine-grained information that encourages deeper visual understanding beyond shallow semantics. A promising direction is introducing annotated captioning / VQA / visual reasoning datasets, which exhibit a stronger alignment of real-world human needs than these captions sourced from the Internet. Concurrent works such as InstructBLIP~\cite{Dai2023InstructBLIP}, Otter~\cite{Li2023Otter}, PaLI-X~\cite{Chen2023PaLI}, and Ying-LM~\cite{Li2023M3IT}, have shown encouraging results of using a collection of vision-language datasets for visual instruction tuning.

However, there exists a significant challenge yet to be resolved in the process of visual instruction tuning. Existing captioning, VQA, and visual reasoning datasets typically provide concise ground truths or answers. However, as human users, we generally prefer AI assistants that can provide ChatGPT-style structured responses, along with optional detailed explanations and elaborations. When using raw annotations for visual instruction tuning, their style would also be learned by the model, even the LLM part is kept frozen and only the connector is tuned. As a result, the InstructBLIP model, the current SoTA model on a wide range of vision-language benchmarks, ranked second to last~\cite{Li2023MIMIC}  in Multi-Modality Arena~\cite{Xu2023LVLM}, a user rating-based evaluation platform of multi-modal LLMs. The model with the lowest Elo rating score is Multimodal-GPT~\cite{Gong2023MultiModal}, which is also tuned with raw annotations. This phenomenon is caused by the additional multi-modal alignment step upon LLM, which thus can be termed as ``\textit{multi-modal alignment tax}'':


\begin{mdframed}[style=coloredbox]
\begin{definition}\textit{
\label{def:tax}
    Multi-modal alignment tax $\Delta P_{\{g,f_\text{LLM}\}}$ is the extra cost of enabling multi-modal perception for LLMs via visual instruction tuning $g$ that maps a text-only $f_\text{LLM}$ to a multimodal LLM, i.e., $g(f_\text{LLM}) \rightarrow f_\text{MLLM}$. The cost is typically reflected as a degradation in task performance that measures model capacity from a certain perspective. Considering a total of $n$ tasks $\{T_1, T_2, ..., T_n\}$ and their corresponding performance measure $P_{T_i}$, the multi-modal alignment tax can be quantified as:$\Delta P_{\{g,f_\text{LLM}\}} = \sum_{i=1}^{n} \left( P_{T_i}(f_\text{LLM}) - P_{T_i}(f_\text{MLLM}) \right)$.
}\end{definition}
\end{mdframed}

The root cause is that: visual representations are fed as soft prompts or prefixes to the LLM, while it is proved that prompt tuning or prefix tuning is able to drastically change the behavior of language models~\cite{He2022Towards}, similar to other parameter-efficient fine-tuning (PEFT) methods such as LoRA~\cite{Hu2022LoRA}. In this paper,  our goal is to prevent LLMs from learning undesired response styles of raw vision-language dataset annotations during visual instruction tuning, thus being a ``\textit{polite}'' multi-modal LLM:

\begin{mdframed}[style=coloredbox]
\begin{definition}
\textit{Polite multi-modal LLMs provide natural and appropriate responses to user queries. Reduction in politeness is a specific instance of multi-modal alignment tax that impacts the model's ability to maintain optimal response styles.}
\end{definition}
\end{mdframed}

To achieve this goal, we introduce a novel method that involves converting these raw responses into natural ones, and we then train the multi-modal LLM using this style-transferred high-quality instruction data, thus mitigating the multi-modal alignment tax on response politeness. As shown in Figure~\ref{fig:overview}, to obtain a rewriter that is capable of transferring the response style, we first distort the ``polite'' version of the response (\textit{e.g.,} GPT-4 generated contents) into an ``impolite'' one, approximating the distribution of existing vision-language dataset annotations. We fine-tune a multi-modal LLM, OpenFlamingo-9B~\cite{Awadalla2023OpenFlamingo}, to learn the reversed mapping (\textit{i.e.,} impolite → polite). Subsequently, we apply the learned model, referred to as ``Polite Flamingo", to rewrite massive annotations in existing vision-language datasets. After carefully filtering out low-quality results and hallucinations, we obtain a high-quality yet large-scale visual instruction tuning dataset PF-1M, and use it to tune a multi-modal LLM. 

We perform a comprehensive evaluation comparing the resulting visual instruction-tuned model, which we called ``Clever Flamingo'', with other multi-modal LLMs, including MiniGPT-4~\cite{Zhu2023MiniGPT}, LLaVA~\cite{Liu2022Visual}, InstructBLIP~\cite{Dai2023InstructBLIP}, and Otter~\cite{Li2023Otter}. In summary, Clever Flamingo outperforms all of these models on detailed image captioning tasks, and only underperforms the InstructBLIP series~\cite{Dai2023InstructBLIP} on VQA tasks (InstructBLIP uses a 3$\times$heavier visual backbone, 8.6$\times$larger pretraining dataset, and +0.6M more instruction samples). For multi-image reasoning tasks, Clever Flamingo outperforms the Otter baseline by a significant margin. In terms of human preference (\textit{i.e.,} politeness), Clever Flamingo only underperforms the LLaVA series~\cite{Liu2022Visual}, which uses purely GPT-4-generated instructions. The contributions of this paper are summarized as follows:

\begin{itemize}
    \item We proposed a novel method to curate raw vision-language datasets into visual instruction tuning data, which enables learning from a wide range of annotated datasets with reduced multi-modal alignment tax.
    \item We constructed a large-scale visual instruction tuning dataset based on response rewriting, and provide empirical solutions to ensure data quality and mitigate hallucinations. 
    \item We further introduced a U-shaped multi-stage visual instruction tuning pipeline and multi-turn augmentations to produce a strong instruction-tuned multi-modal LLM efficiently.
    \item We performed comprehensive evaluations in terms of both multi-modal understanding and response politeness using automated evaluators, whose reliability is verified by human evaluations. 
\end{itemize}

\begin{figure}
    \centering
    \includegraphics[width=1\linewidth]{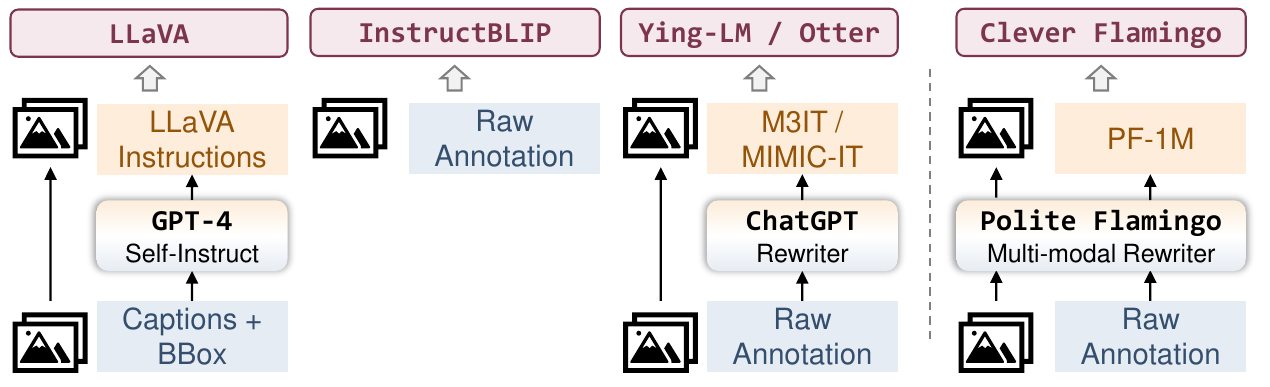}
    \caption{\textbf{Comparison of different visual instruction tuning methods}. \textbf{LLaVA}~\cite{Liu2022Visual} performs multi-modal self-instruct~\cite{Wang2022Self} using GPT-4, which has high API cost and limited visual groundedness; \textbf{InstructBLIP}~\cite{Dai2023InstructBLIP} directly uses learn raw annotations, and thus suffer from multi-modal alignment tax; \textbf{M$^3$IT}~\cite{Li2023M3IT} and \textbf{MIMIC-IT}~\cite{Li2023MIMIC} employed ChatGPT-based rewriters, while we train a \textbf{Polite Flamingo} to rewrite responses, which enjoys advantages of 1) multi-modality, 2) scalability, and 3) diversity.}
    \label{fig:compare_tuning_data}
\end{figure}

\section{Related Works}

\textbf{Visual instruction tuning for multi-modal LLM}. Research on enabling visual perception for powerful but blind LLMs attracted widespread attention recently~\cite{Yin2023Survey}. The most straightforward methodology is to integrate image captioning experts via prompt engineering (\textit{e.g.,} Socratic Models~\cite{Zeng2022Socratic}, HuggingGPT~\cite{Shen2023HuggingGPT}, MM-REACT~\cite{Yang2023MM}). However, this is inefficient due to the low bandwidth of natural language communication: given the diversity of real-world visual tasks, describing all of the potential task-relevant information within a single image requires a huge amount of language tokens. Therefore, many efforts opt to connect compact latent visual representations through a dense connector by visual instruction tuning, such as MiniGPT-4~\cite{Zhu2023MiniGPT}, LLaVA~\cite{Liu2022Visual}, Multimodal-GPT~\cite{Gong2023MultiModal}, LLaMA-Adapter~\cite{Zhang2023LLaMA}, Otter~\cite{Li2023Otter}, mPLUG-Owl~\cite{Ye2023mPLUG}, InstructBLIP~\cite{Dai2023InstructBLIP}. These models use linear projectors or perceivers as the connector between visual models and LLM, thus having a much larger information bandwidth compared to those prompt-based natural language communications.

\textbf{Data for visual instruction tuning}. However, what data is optimal for training these connectors to ensure that they propagate visual information faithfully is unclear. Existing attempts include generating self-instruct~\cite{Wang2022Self} data (\textit{i.e.,} LLaVA~\cite{Liu2022Visual}), using image-text captioning datasets (\textit{e.g.,} COCO~\cite{Chen2015Microsoft}, SBU~\cite{Ordonez2011Im2Text}, CC-3M~\cite{Sharma2018Conceptual}), and unifying downstream vision-language datasets (\textit{e.g.,} VQA and visual reasoning datasets). Although GPT-4 generated LLaVA dataset enjoy very high quality, its scale remains insufficient, and it could not encourage fine-grained vision-language alignment, as it does not ``make V in VQA matter''~\cite{Goyal2017Making}. On the other hand, using captioning datasets only would result in degenerated QA capabilities, as a soft prompt that encourages image captioning is implicitly learned by the connector, then the model would prefer to give an image caption even if the instruction asks it to answer a certain question. 
 
\textbf{Multi-modal alignment tax}. Therefore, many efforts have been focused on utilizing downstream vision-language datasets, including Multimodal-GPT~\cite{Gong2023MultiModal}, Otter~\cite{Li2023Otter}, InstructBLIP~\cite{Dai2023InstructBLIP}, M$^3$IT~\cite{Li2023M3IT}, LAMM~\cite{Yin2023LAMM}. Unfortunately, the multi-modal alignment tax (Definition~\ref{def:tax}) becomes a serious side effect that destroys the response formatting ability of the resulting multi-modal LLMs. To avoid such cost, the earliest work Multimodal-GPT~\cite{Gong2023MultiModal} simply removed vision-language datasets that contain short answers. InstructBLIP~\cite{Dai2023InstructBLIP} adds additional prompts such as ``provide your answer as short as possible'' to the instruction, but still could not mitigate the short answer bias due to the imbalance of response style -- most responses in the training data are very short so the model just ignores these additional prompts.

\textbf{ChatGPT-based text-only rewriter}. Another attempt to mitigate the multi-modal alignment tax is to use ChatGPT to rewrite the short answer, as adopted in concurrent works M$^3$IT~\cite{Li2023M3IT} and MIMIC-IT~\cite{Li2023MIMIC}. We compare our method with them in Figure~\ref{fig:compare_tuning_data}. Since our Polite Flamingo is a \textit{multi-modal} rewriter, it can fuse visual perception with text semantics to rewrite, as opposed to these ChatGPT-based blind models that can only rely on the answer information. Polite Flamingo is also much lighter, cheaper, and does not require any API cost, leading to better scalability\footnote{Polite Flamingo is based on LLaMA-7B and can be run on consumer GPUs. BF-16 inference of Polite Flamingo roughly takes 18 GB GPU memory.}. Moreover, Polite Flamingo is specially trained on 255k diverse rewriting examples, while ChatGPT can only perform zero-shot or few-shot rewriting. As an example of its limitation, M$^3$IT~\cite{Li2023M3IT} used a single in-context rewriting demonstration to prompt ChatGPT, which resulted in limited diversity -- 96\% rewritten samples within its A-OKVQA subset have the sentence pattern of ``\texttt{\{rational\}, so the answer is \{answer\}}". Finally, our work also shares some similarities with FuseCap~\cite{Rotstein2023FuseCap}
and LaCLIP~\cite{Fan2023Improving} and RemoteCILP~\cite{Liu2023RemoteCLIP} that generate/rewrite image captions to train vision language models.

\section{Polite Flamingo: a Multi-modal Instruction Response Rewriter}

\begin{figure}[]
    \centering
    \includegraphics[width=1\linewidth]{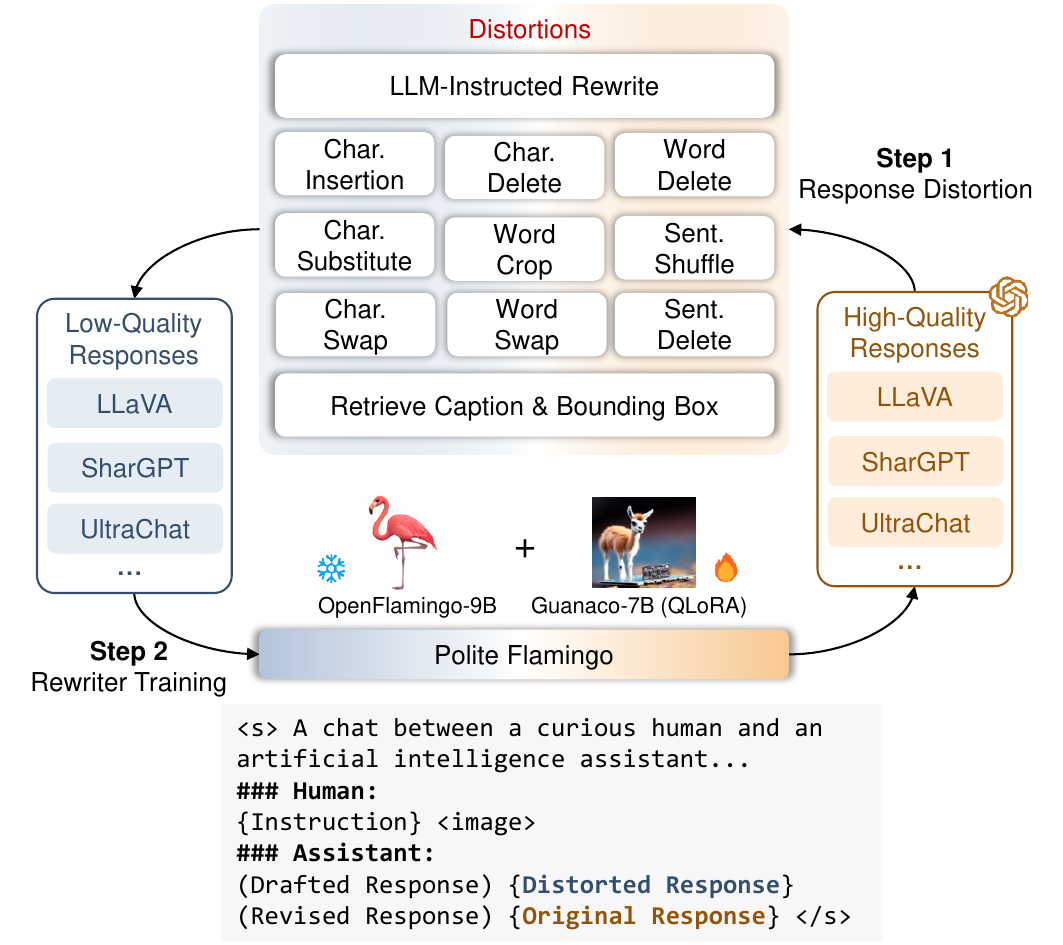}
    \caption{\textbf{Training pipeline of training Polite Flamingo.} We distort original high-quality responses into the corresponding low-quality version, then train a multi-modal LLM to predict the original response. This model is then used to rewrite raw annotations of a wide range of vision-language datasets and derive a PF-1M dataset for visual instruction tuning.}
    \label{fig:polite_flamingo_pipeline}
\end{figure}

To learn a rewriter for raw annotations of vision-language datasets, the most straightforward way could be to train a model to directly predict a ``polite'' version from the corresponding raw annotations. Unfortunately, careful annotation of such translations is highly expensive and hard to scale. To overcome this limitation, we design a surrogate task that trains the rewriter to learn the style from existing high-quality instruction data, such as the LLaVA self-instruct dataset~\cite{Liu2023Visual}. Specifically, we first transfer the style of these high-quality responses into low-quality ones, approximating the distribution of the raw annotations in the vision-language dataset that needs to be rewritten. Then, we train the model to reconstruct the original high-quality response from given distortions, as shown in Figure~\ref{fig:polite_flamingo_pipeline}.

Our methodology is inspired by denoising AutoEncoder-style image enhancement models. These systems automatically introduce distortions, such as random noise or down-sampling, to the original images, and then the model is trained to reconstruct the original images. The resulting model can then be applied to image denoising or super-resolution. The key assumption of these image enhancement models, as well as our Polite Flamingo is that the distortion module should produce samples \textit{i.i.d.} to the input samples during inference (\textit{i.e.,} noise/low-resolution images, or raw annotations) so that the train-test domain divergence is small and these denoising AutoEncoders can generalize well.

\subsection{Response Distortion}
\label{sec:response_distortion}
\begin{figure*}
    \centering
    \includegraphics[width=\linewidth]{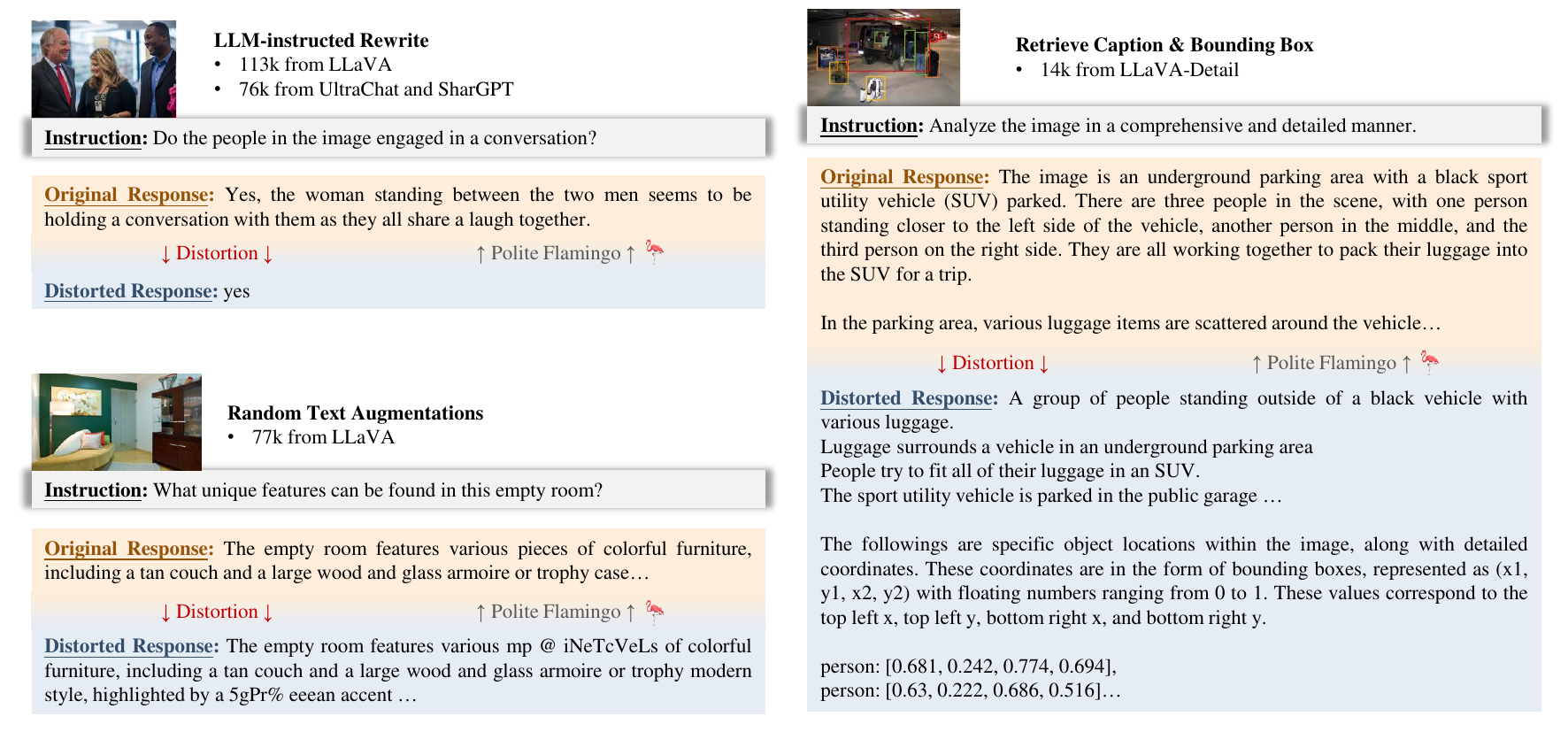}
    \caption{\textbf{Examples of our three response distortion strategies.} We transfer the style of LLM-generated high-quality instruction responses into an ``impolite'' version, approximating the distribution of raw vision-language dataset annotations. The converted data is used to train Polite Flamingo.}
    \label{fig:distortion_examples}
\end{figure*}

To approximate the distribution of raw vision-language dataset annotations that would be used for Polite Flamingo inference, we develop the following three strategies for response distortion. Resulting examples are shown in Figure~\ref{fig:distortion_examples}.

\begin{itemize}
    \item \textbf{LLM-instructed Distortion}. Representative patterns of raw annotations include short answers (\textit{e.g.,} VQA-v2~\cite{Goyal2017Making}), lacking punctuation or capitalization (\textit{e.g.,} MS-COCO Captions~\cite{Chen2015Microsoft}), not being coherent (\textit{e.g.,} A-OKVQA~\cite{Schwenk2022OKVQA}), etc., and we prompt an LLM (Guanaco~\cite{Dettmers2023QLoRA}\footnote{We used the QLoRA-based Guanaco language model~\cite{Dettmers2023QLoRA}, known for its superior performance (33B version, which has an average win rate of 97.8\% against ChatGPT evaluated by GPT-4).}) to produce responses similar to these patterns. For each sample, we append another round of conversation, asking the model to transfer the original response into a ``impolite'' one. Furthermore, we randomly sample a distortion command from a pool containing a total of 24 alternatives and add it to the prompt with a probability of 50\%. The distortion choices, which aim to further mimic the style of raw annotations, include capitalization modifications, inserting repetitions, using incorrect tenses, removing formatting, adding irrelevant information, etc. See Table~\ref{tab:llm_instructed_distortion_prompt} in the appendix for the detailed prompt structure.

    \item \textbf{Random Text Augmentations}. This distortion is much cheaper compared to LLM-based distortion, and we introduce it to further increase the diversity of the Polite Flamingo training set. Specifically, We use the \texttt{NLPAUG}\footnote{\url{https://github.com/makcedward/nlpaug}} library to perform character-level, word-level, and sentence-level text augmentation.
    Every level of augmentation is applied with a probability of 50\%.

    \item \textbf{Retrieve Captions \& Bounding Boxes}. In the LLaVA dataset~\cite{Liu2023Visual}, GPT-4 is used to produce high-quality detailed captions for visual instruction tuning, given five captions and all bounding box annotations of each image. However, possibly due to the high API cost, there are only 23k samples of such detailed descriptions. Here we would like to distill such capability into the Polite Flamingo, and extrapolate it into the remaining MS-COCO samples, as well as other datasets with multiple captions (\textit{e.g.,} Flicker-30k) or bounding box annotations (detection datasets). We retrieve the original captions and object bounding boxes in the \texttt{LLaVA-detailed-23k} dataset and use them as the distorted version with respect to the original detailed descriptions. We also insert the description of ``The followings are specific object locations...'' which was used for prompting GPT-4, to help Polite Flamingo understand bounding box annotations.
\end{itemize}

\subsection{Source Datasets}
\label{sec:polite_flamingo_source_datasets}

When selecting the source datasets for training Polite Flamingo, we take into consideration the following three criteria. \textbf{1) Politeness}: The source datasets chosen should contain responses with a desired level of politeness. These responses will be directly learned by Polite Flamingo and subsequently transferred to the final model.  \textbf{2) Multi-modality}: It is important for Polite Flamingo to leverage complementary visual information during the process of response rewriting. We expect it can provide necessary explanations for those short answers to ensure comprehensive and informative responses. \textbf{3) Diversity}: The training set must be sufficiently large to prevent the LLM-based Polite Flamingo from overfitting to specific patterns. According to the above criteria, we select three datasets to construct the training data for Polite Flamingo:

\begin{enumerate}
    \item \textbf{LLaVA instructions~\cite{Liu2023Visual}}: a multi-modal self-instruct dataset based on GPT-4, which is currently the only available LLM-generated multi-modal visual instruction tuning dataset. In this study, we assume that ChatGPT/GPT-4 produces responses that are considered satisfactory in terms of style\footnote{Since our methodology is data-driven, it is not limited to this particular style. Polite Flamingo can easily incorporate and adapt to other styles if we have access to sufficient high-quality data from other sources.}. Therefore, this dataset satisfies the criteria of both politeness and multi-modality.
    \item \textbf{UltraChat~\cite{Ding2023Enhancing}}: a large-scale text-only instruction dataset consisting of dialogues between two ChatGPT turbo APIs. Since the LLaVA instructions dataset contains only 117k data points, we select this dataset to compensate for the limited data diversity. UltraChat is generated by ChatGPT and has undergone post-processing and careful filtering~\cite{Ding2023Enhancing}, so we assume it provides satisfactory politeness.
    \item \textbf{ShareGPT}: a dataset of conversations with ChatGPT that is shared by users and was used to train the Vicuna model. This dataset contains model responses to real-world user queries, resulting in good diversity. ShareGPT is also considered to be of high quality, as the resulting models (Vicuna) have shown superior performance~\cite{Zheng2023Judging}.
\end{enumerate}


\subsection{Training a Rewritter}

We gathered a total of 255k samples to train the Polite Flamingo. We initialize the model from OpenFlamingo-9B~\cite{Awadalla2023OpenFlamingo}, and insert a LoRA~\cite{Hu2022LoRA} adapter (initialized from the QLoRA of Guanaco-7B~\cite{Dettmers2023QLoRA}) into its LLaMA-7B~\cite{Touvron2023LLaMA} language model. We tune the LoRA weights only, and keep other parameters (\textit{i.e.,} language model, ViT, perceiver, X-ATTN layers~\cite{Alayrac2022Flamingo}) frozen to prevent overfitting. As shown in Figure~\ref{fig:polite_flamingo_pipeline}, we provide the instruction, image, and distorted response to the Polite Flamingo, and ask it to predict the original response. Language modeling loss is only applied to the tokens corresponding to the original response.

\section{Scale Up Visual Instruction Tuning with Polite Flamingo}


\subsection{Source Datasets}
\label{sec:source_datasets_PF_inference}

To scale up the vision-language instruction tuning data thus improving the visual understanding capability of the multi-modal LLM, we leverage the trained Polite Flamingo to rewrite the raw annotations of numerous vision-language datasets into polite responses. Similar to several concurrent works~\cite{Dai2023InstructBLIP,Li2023M3IT, Li2023MIMIC}, we standardize them into a unified instruction-response format. The adopted datasets can be roughly divided into two main groups: captioning datasets, which task the model with providing detailed descriptions of image content, and VQA datasets, which require the model to accurately answer specific queries. We adopted a total of 37 datasets, including MS-COCO~\cite{Chen2015Microsoft}, Flickr-30k~\cite{Young2014image}, TextCaps~\cite{Sidorov2020TextCaps}, Image2Paragraph~\cite{Krause2017Hierarchical}, CC-3M~\cite{Sharma2018Conceptual}, ELEVATER-IC~\cite{Li2022ELEVATER}, Spot-the-Diff~\cite{Jhamtani2018Learning}, Image-editing-requests~\cite{Tan2019Expressing}, RefCOCOg~\cite{Yu2016Modeling}, A-OKVQA~\cite{Schwenk2022OKVQA}, VQA-E~\cite{Li2018VQA}, ScienceQA~\cite{Lu2022Learn}, VQA-v2~\cite{Goyal2017Making}, GQA~\cite{Hudson2019GQA}, OCR-VQA~\cite{Mishra2019OCR}, PointQA~\cite{Mani2020Point}, etc. We summarized detailed information in Section~\ref{sec:Clever Flamingo Training Data} and Table~\ref{tab:PF-1M} in the appendix.

\subsection{Filtering Strategies}
\label{sec:filtering}

Our rewriter, Polite Flamingo, is based on LLaMA-7B~\cite{Touvron2023LLaMA}, which is a relatively small language model. Through empirical observation, we have identified that Polite Flamingo is not a flawless response rewriter. It occasionally leaves the answer unchanged, produces repetitive patterns, or even changes the original answer and introduces hallucinated content. 
We design an automatic filtering pipeline to mitigate these problems and guarantee the quality of visual instruction tuning data. We use several rule-based filters, and several newly introduced model-based filters to measure the semantics of rewritten response, including a Semantic Textual Similarity (STS) model-based filter, a Natural Language Inference (NLI) model-based filter, and a CLIPScore-based hallucination filter. See Appendix~\ref{sec:filters} for implementation details.

\subsection{U-shaped Multi-stage Visual Instruction Tuning}
\label{sec:U-shape}

We first leverage the Polite Flamingo to rewrite the response of source datasets (Section~\ref{sec:source_datasets_PF_inference}), obtaining 1.17M samples. After filtering, 0.97M samples remained, which we refer to as the PF-1M dataset. In addition to PF-1M, we also adopt several high-quality text-only instruction datasets, since our base model OpenFlamingo-9B is based on the vanilla LLaMA-7B which is not instruction-tuned. Recent studies have shown that data quality is of vital importance during instruction tuning. Motivated by this, we consider the following datasets: UltraChat~\cite{Ding2023Enhancing}, ShareGPT, OASST-1~\cite{Koepf2023OpenAssistant}, Alpaca-GPT-4~\cite{Peng2023Instruction}, GPTeacher, and InstructionWild~\cite{Xue2023Instruction}. Together with PF-1M and LLaVA-instruction-177k, we have a total of 1.5M instruction data.

However, the samples in this dataset collection provide benefits to the model from very different perspectives. Text-only instructions enable the model to comprehend human requests and generate helpful responses in a proper style, while PF-1M data primarily facilitate the model in improving precise visual perception. To enhance training efficiency, we propose a U-shaped visual instruction tuning approach that encompasses three stages:

    \textbf{Stage 1} focuses on improving the instruction-following ability of the model by tuning only the language model (with LoRA). We utilize a total of 0.77M samples, which include all text-only instructions, LLaVA instructions, and 10\% samples (97k) from PF-1M, and trained the model for a single epoch. The model is trained with a large context window of 1024 tokens. 
    \textbf{Stage 2} shifts to improving the visual understanding capability of the model. We freeze the LoRA adapter and exclusively tune the connector using the entire PF-1M dataset. To enhance training efficiency, we use a smaller context window of 196 tokens.
    \textbf{Stage 3} uses the same setting as Stage 1, but we adjust the learning rate to 10$\times$ lower. The objective of Stage 3 is to fine-tune the model to recover the optimal politeness of the responses. This adjustment is necessary as the PF-1M dataset used in Stage 2 is generated by a 7B language model, which has lower quality than larger LLM-generated text-only instructions.

\label{sec:qualitative_evaluation}
\begin{figure*}
    \centering
    \includegraphics[width=0.94\linewidth]{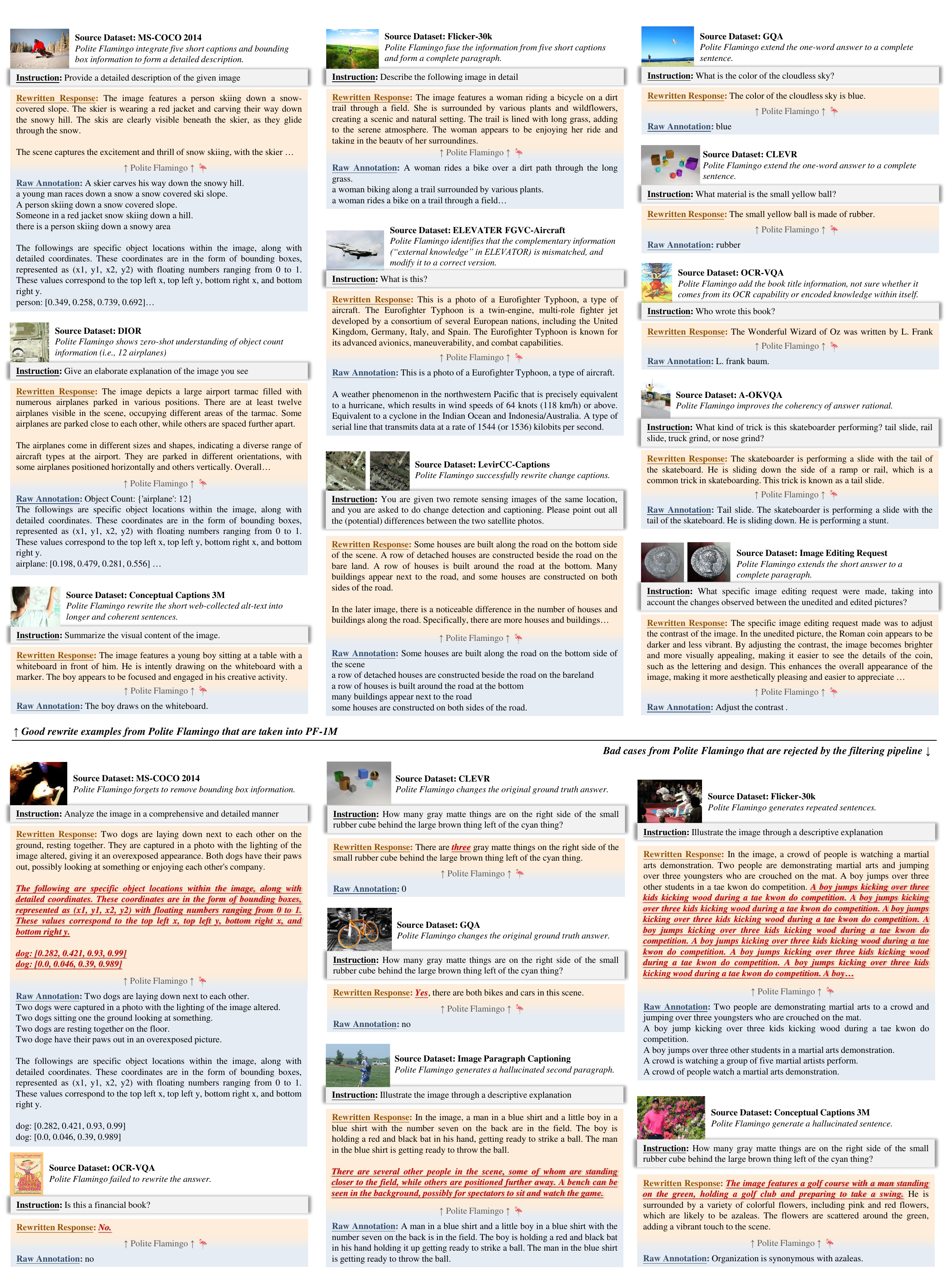}
    \caption{\textbf{Representative examples of Polite Flamingo-based response rewriting.} We show both good examples that are taken into PF-1M (upper), and bad cases that are rejected by our filtering pipeline (bottom).}
    \label{fig:rewrite_examples}
\end{figure*}

\subsection{Multi-turn Augmentation}
\label{sec:multi-turn}
Given the diversity of instruction data, the length of each sample varies a lot. When using a large context window, short instruction samples would append many \texttt{<PAD>} tokens and waste a lot of computation. To address this, we introduce multi-turn augmentation, which involves randomly selecting instruction samples and concatenating them to form a multi-turn conversation. In this augmentation scheme, only the tokens corresponding to the response in each turn are considered when calculating the language modeling loss. This multi-turn also encourages the model to attend to the correct image for multi-turn multi-image conversations.

\section{Evaluations}

\subsection{How Does Polite Flamingo Rewrite the Response?}

\subsubsection{Qualitative Evaluation}

First, we present a qualitative analysis of Polite Flamingo's rewriting. In Figure~\ref{fig:rewrite_examples}, we show representative examples of both good (upper) and bad (bottom) cases, and note how Polite Flamingo rewrites examples as expected and how it makes mistakes. Overall, Polite Flamingo successfully converts raw annotations into polite, rich, and coherent responses. From various examples, it is observed that it is capable of 1) integrating information from multiple captions and/or bounding boxes, 2) improving response coherency, and 3) generating complete sentences/paragraphs from short annotations, etc.

\textbf{Good Cases.} One interesting example is shown in the center of the upper half -- the ``Eurofighter Typhoon'' from ELEVATER's FGVC-Aircraft dataset. The source dataset provides external knowledge retrieved from Wikipedia, WordNet, and GPT-3, as knowledge augmentations. However, in this example, the original external knowledge is mismatched with the image due to word ambiguity (a type of aircraft vs. a climate concept). As Polite Flamingo is a multi-modal LLM that can observe both image and text, it recognized this mismatch and modified it to the correct version. Another example is shown on the right side of the Typhoon example (from OCR-VQA~\cite{Mishra2019OCR} dataset), in which the Polite Flamingo added the book title information to its rewritten answer. These examples illustrate the advantage of Polite Flamingo-based response rewriting in comparison with those ChatGPT-based ones (\textit{e.g.,} in MIMIC-IT~\cite{Li2023MIMIC}, M3IT~\cite{Li2023M3IT}, FuseCap~\cite{Rotstein2023FuseCap}, etc.). The multi-modality understanding ability of Polite Flamingo enables it to have a more comprehensive understanding of the instruction-response sample than the text-only rewriters. 

\textbf{Bad Cases.} However, compared to ChatGPT-based rewriters, a major drawback of Polite Flamingo is its reliability -- Polite Flamingo still makes some silly mistakes. In the bottom half of Figure~\ref{fig:rewrite_examples}, we show some representative examples of low-quality rewriting. Despite simple mistakes such as forgetting to generate \texttt{<EOS>} token thus producing endless repetitions, notable issues include changing the ground truth answer or adding hallucinated contents. It seems that sometimes Polite Flamingo prefers to believe its own visual perception rather than the provided ground truth, and its visual perception is not always accurate -- possibly because the base model of Polite Flamingo, the OpenFlamingo-9B, is only trained on 15M image-text data thus produce less comprehensive visual representation alignment. These examples also demonstrate the necessity of post-processing and filtering.

\begin{table*}[]
\caption{\textbf{Performance comparison of different multi-modal LLMs}. We use Rouge-L as the metric for detailed image description tasks (MS-COCO, TextCaps, and Image-to-paragraph), and we use an NLI-based evaluator for VQA datasets (OK-VQA, Visual-Spatial Reasoning, and Grid-3D). \textbf{\textcolor[HTML]{809EC2}{Blue numbers}} are results on unseen datasets (\textit{i.e.,} zero-shot), and \textbf{black numbers} are results on unseen samples (\textit{i.e.,} validation split of datasets seen during training). The bottom row ($\pm\Delta$) compares Clever Flamingo with Otter, which uses the same OpenFlamingo-9B as the base model.}
\label{tab:benchmark}
\centering 
\resizebox{\textwidth}{!}{%
\begin{tabular}{ccccccccccc}
\hline
 &
   &
   &
   &
   &
  \multicolumn{3}{c}{\textbf{Detailed   Image Description}} &
  \multicolumn{3}{c}{\textbf{Visual   Question Answering}} 
  \\ 
  \cline{6-11} 
\multirow{-2}{*}{\textbf{Method}} &
  \multirow{-2}{*}{\textbf{\#Instructions}} &
  \multirow{-2}{*}{\textbf{\begin{tabular}[c]{@{}c@{}}Visual\\      (\#Params)\end{tabular}}} &
  \multirow{-2}{*}{\textbf{\begin{tabular}[c]{@{}c@{}}Connector\\      (\#Samples)\end{tabular}}} &
  \multirow{-2}{*}{\textbf{\begin{tabular}[c]{@{}c@{}}LLM\\      (\#Params)\end{tabular}}} &
  \textbf{COCO} &
  \textbf{TextCaps} &
  \textbf{Img2P} &
  \textbf{OK-VQA} &
  \textbf{VSR} &
  \textbf{Grid-3D}
  \\ \hline
 &
   &
   &
   &
  7B &
  14.4 &
  {\color[HTML]{809EC2} 15.5} &
  {\color[HTML]{809EC2} 14.7} &
  {\color[HTML]{809EC2} 10.4} &
  {\color[HTML]{809EC2} 14.0} &
  {\color[HTML]{809EC2} 19.0} 
  \\
\multirow{-2}{*}{MiniGPT-4} &
  \multirow{-2}{*}{3.5k} &
  \multirow{-2}{*}{ViT-g (1.0B)} &
  \multirow{-2}{*}{Linear (5M)} &
  13B &
  23.1 &
  {\color[HTML]{809EC2} 19.2} &
  {\color[HTML]{809EC2} 23.7} &
  {\color[HTML]{809EC2} 23.8} &
  {\color[HTML]{809EC2} 24.6} &
  {\color[HTML]{809EC2} 20.0} 
  \\ \hline
 &
   &
   &
   &
  7B &
  23.8 &
  {\color[HTML]{809EC2} 21.1} &
  {\color[HTML]{809EC2} 23.6} &
  {\color[HTML]{809EC2} 32.1} &
  {\color[HTML]{809EC2} 36.1} &
  {\color[HTML]{809EC2} 20.8}
  \\
\multirow{-2}{*}{LLaVA} &
  \multirow{-2}{*}{177k} &
  \multirow{-2}{*}{ViT-L (0.3B)} &
  \multirow{-2}{*}{Linear (595k)} &
  13B &
  23.1 &
  {\color[HTML]{809EC2} 20.7} &
  {\color[HTML]{809EC2} 23.2} &
  {\color[HTML]{809EC2} 30.9} &
  {\color[HTML]{809EC2} 34.1} &
  {\color[HTML]{809EC2} 22.5} 
  \\ \hline
 &
   &
   &
   &
  7B &
  23.7 &
  22.2 &
  {\color[HTML]{809EC2} 22.2} &
  51.5 &
  {\color[HTML]{809EC2} 48.5} &
  {\color[HTML]{809EC2} 28.9} 
  \\
\multirow{-2}{*}{\begin{tabular}[c]{@{}c@{}}InstructBLIP\\      (Vicuna)\end{tabular}} &
  \multirow{-2}{*}{1.6M} &
  \multirow{-2}{*}{ViT-g   (1.0B)} &
  \multirow{-2}{*}{BLIP-2 (129M)} &
  13B &
  23.5 &
  19.7 &
  {\color[HTML]{809EC2} 22.1} &
  {\ul \textbf{52.2}} &
  {\color[HTML]{809EC2} {\ul \textbf{48.9}}} &
  {\color[HTML]{809EC2} 27.5} 
  \\ \hline
Otter &
  2.8M &
   &
   &
  7B &
  22.6 &
  {\color[HTML]{809EC2} 19.7} &
  {\color[HTML]{809EC2} 22.4} &
  {\color[HTML]{809EC2} 28.7} &
  {\color[HTML]{809EC2} 28.7} &
  {\color[HTML]{809EC2} 13.5} 
  \\
\textbf{Clever   Flamingo}  &
  1.0M &
  \multirow{-2}{*}{ViT-L   (0.3B)} &
  \multirow{-2}{*}{OpenFlamingo-9B   (15M)} &
  7B &
  {\ul \textbf{24.3}} &
  {\ul \textbf{24.1}} &
  {\ul \textbf{24.7}} &
  {\color[HTML]{809EC2} 43.3} &
  {\color[HTML]{809EC2} 43.6} &
  {\color[HTML]{809EC2} {\ul \textbf{29.0}}} 
  \\
\rowcolor[HTML]{DAE8FC} 
$\pm\Delta$ &
  {\color[HTML]{CB0000} \textbf{-1.8M}} &
  - &
  - &
  - &
  {\color[HTML]{00B050} \textbf{ +1.7}} &
  {\color[HTML]{00B050} \textbf{+4.4}} &
  {\color[HTML]{00B050} \textbf{+2.3}} &
  {\color[HTML]{00B050} \textbf{+14.6}} &
  {\color[HTML]{00B050} \textbf{+14.9}} &
  {\color[HTML]{00B050} \textbf{+15.5}} 
  \\ \bottomrule
\end{tabular}%
}
\end{table*}

\subsubsection{Quantitative Evaluation}

\begin{figure}
    \centering
    \includegraphics[width=\linewidth]{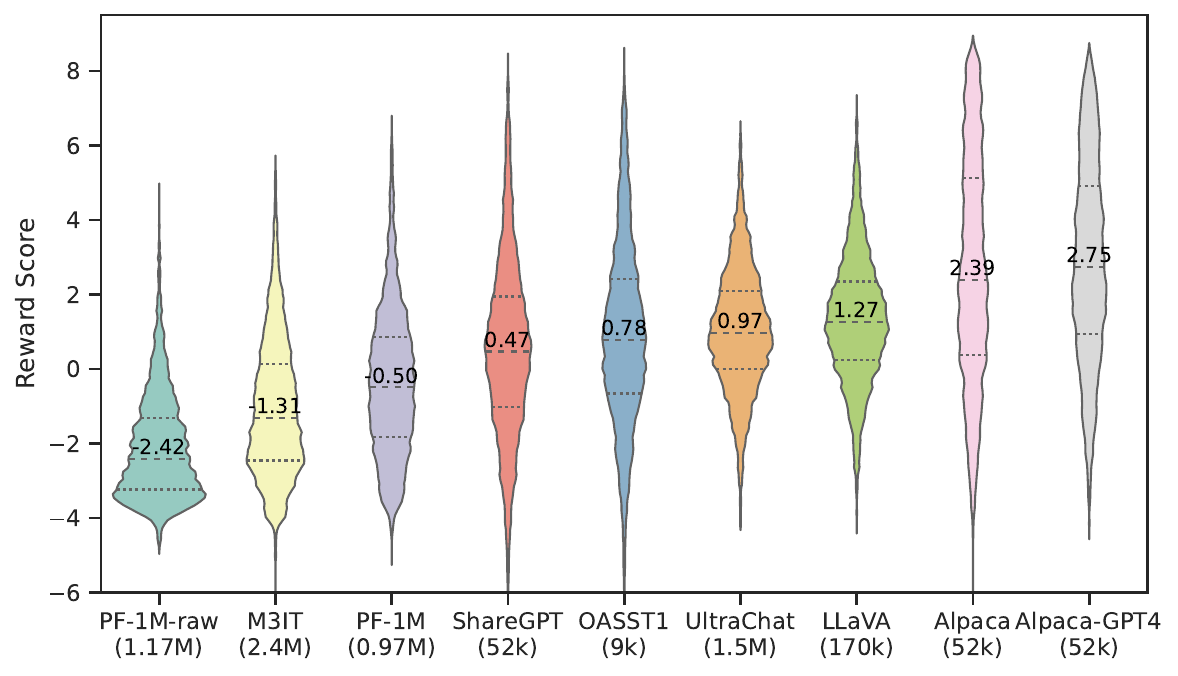}
    \caption{\textbf{Distribution of absolute reward model score of various instruction tuning datasets}. The median and quartile are also marked by dotted lines. Polite Flamingo boost the ``politeness'' of raw dataset annotations (leftmost) significantly.}
    \label{fig:dataset_distribution}
\end{figure}

In addition to the above examples, we analyze the improvement of ``politeness'' through a quantitative evaluation. We assume that a reward model which is trained on human-labeled user preference data is able to provide an estimation of politeness\footnote{Reward model: \url{https://huggingface.co/OpenAssistant/reward-model-deberta-v3-large-v2}}. In Figure~\ref{fig:dataset_distribution}, we plot the distribution of the scores of the reward model on a wide range of popular instruction tuning datasets\footnote{10k samples are randomly drawn from each dataset. At the time of writing, another concurrent visual instruction tuning dataset MIMIC-IT~\cite{Li2023MIMIC} (which is used to train Otter) is not fully available.}.  It shows that Polite Flamingo significantly boosts the politeness of raw dataset annotations (from -2.42 to -0.50), and the resulting PF-1M outperforms the recently proposed large-scale instruction tuning dataset M$^3$IT~\cite{Li2023M3IT} by a notable margin. Unfortunately, PF-1M cannot match those datasets produced by much larger LLM, especially those generated by GPT-4 (\textit{i.e.,} LLaVA~\cite{Liu2023Visual} and Alpaca-GPT-4~\cite{Peng2023Instruction}). But on the other hand, PF-1M is approximately 6$\times$ larger than the LLaVA dataset, and many LLaVA instructions are QA conversations under the theme of the image. In comparison, the PF-1M dataset is derived from annotated vision-language dataset and involves challenging samples that encourage fine-grained visual understanding.

\subsection{Comparing Clever Flamingo with Existing Multi-modal LLMs}

Now we turn to verify the performance of the Clever Flamingo trained with PF-1M, and compare it with other multi-modal LLMs. We focus on answering the following questions: 1) how well does it perform on vision-language tasks, 2) how does it generalize to unseen datasets, and 3) whether it produces human-preferred responses (\textit{i.e.,} being polite). We first compare it with other models on image captioning and VQA tasks (Section~\ref{sec:single_image}), then we present the evaluation of multi-image reasoning tasks (Section~\ref{sec:multi_image}), and finally, we analyze the politeness of these multi-modal LLMs (Section~\ref{sec:politeness_evaluation}).

\begin{figure*}
    \centering
    \includegraphics[width=1\linewidth]{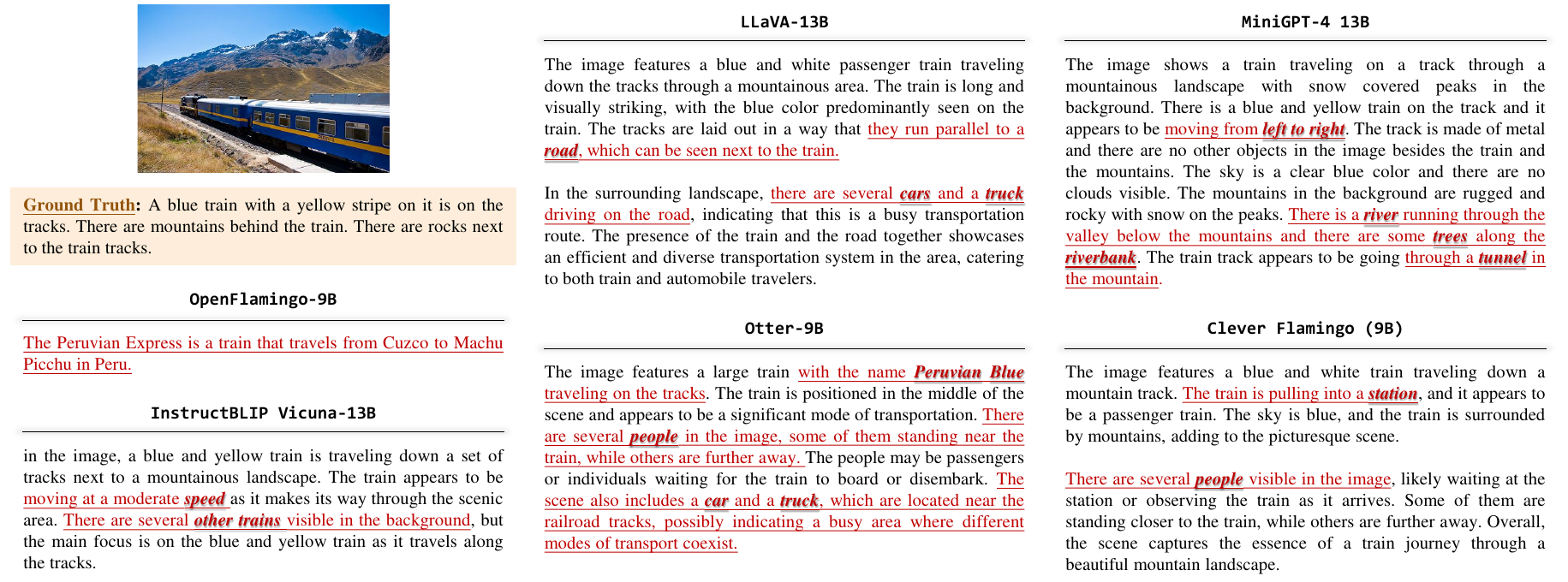}
    \caption{\textbf{All existing multi-modal LLMs exhibit severe hallucination problems.} We show a random testing example from the Image2Paragraph~\cite{Krause2017Hierarchical}. The hallucinated contents are marked with \textcolor[HTML]{cb0000}{\textbf{red}}.}
    \label{fig:hallucinate_example}
\end{figure*}

\subsubsection{Image Captioning and VQA}
\label{sec:single_image}

Table~\ref{tab:benchmark} summarized the evaluation results comparing Clever Flamingo with other multi-modal LLMs on detailed image captioning (MS-COCO~\cite{Chen2015Microsoft}, TextCaps~\cite{Sidorov2020TextCaps}, and Image2Paragraph~\cite{Krause2017Hierarchical}) and visual question answering (OK-VQA~\cite{Marino2019OK}, Visual-Spatial Reasoning~\cite{Liu2022Visual}, Grid-3D~\cite{Lee2022What}). We use Rouge-L as the metric for captioning datasets and use an NLI model-based automated evaluator for VQA datasets (see Section~\ref{sec:evaluate_evaluators} for more details). As our work is concurrent with InstructBLIP~\cite{Dai2023InstructBLIP} and Otter~\cite{Li2023Otter}, the dataset splitting (\textit{i.e.,} assignments of held-in training datasets and held-out unseen testing datasets) is not fully aligned. We marked the held-in datasets with \textbf{black} and marked the held-out datasets with \textcolor[HTML]{809EC2}{\textbf{blue}}. 

In summary, Clever Flamingo outperforms other counterparts on all three detailed image description datasets and the Grid-3D dataset, and only underperforms the InstructBLIP series on OK-VQA and VSR. Importantly, the settings (\textit{e.g.,} the base model and training data amount) of these comparisons are not aligned. For InstructBLIP, a BERT-based Q-Former is firstly trained with  BILP-generated and filtered 129M samples for two stages (about 3-4 epochs), then the model is instruction-tuned on a 1.6M collection of downstream data. In comparison, our Clever Flamingo, as well as the Otter model, is tuned from OpenFlamingo-9B, which uses a 3$\times$smaller visual encoder, a lighter perceiver as the connector, and much less pre-training image-text data (15M) and training steps (single epoch)\footnote{\url{https://laion.ai/blog/open-flamingo/}}~\cite{Awadalla2023OpenFlamingo}. When come to a fair comparison between Clever Flamingo and Otter (despite instruction data, Clever Flamingo uses 1.8M fewer data), our model outperforms Otter on every dataset, both held-in and held-out, by a significant margin.

\textbf{Hallucination Problem}. Although Clever Flamingo yields notable improvement in image captioning tasks, it still exhibits severe object hallucination problems~\cite{Ji2023Survey,Li2023Evaluating,Dai2023Plausible}, the same as other existing multi-modal LLMs. In  Figure~\ref{fig:hallucinate_example},  we prompt existing multi-modal LLMs with the instruction ``Give an elaborate explanation of the image you see'', using a random testing sample\footnote{Not cherry/lemon-picked -- it is the first image in our sampled validation set.} from the Image2Paragraph~\cite{Krause2017Hierarchical} dataset. As marked with red, all of the compared models hallucinated non-exist objects, such as road, cars, trucks, people, river, trees, tunnel, station, etc. This is a significant limitation faced by existing multi-modal LLMs, preventing them to be actually deployed in the real world. We also find that it is difficult to quantitatively verify the correctness of generation beyond object appearance (\textit{e.g.,} ``this is a scenic area'', ``the train is visually striking'', ``beautiful mountain landscape'', ``this is a busy transportation route'', etc.), as we lack a dataset with rich fine-grained annotations of all information that can be inferred from the image.

\subsubsection{Multi-image Reasoning}
\label{sec:multi_image}

Now we analyze the performance on multi-image reasoning tasks. We compare Clever Flamingo with Otter~\cite{Li2023Otter}, which is also tuned from OpenFlamingo-9B -- the only currently publicly available base multi-modal LLM that can process interleaved image-text data. The following three datasets are used for evaluation: 1) Spot-the-diff~\cite{Jhamtani2018Learning}, a change captioning dataset for surveillance camera imagery, 2) Image-editing-requests~\cite{Tan2019Expressing}, which requires the model to infer image editing requests (\textit{e.g,} Photoshop editing) given image pairs, and 3) Natural Language Visual Reasoning-2 (NVLR2)~\cite{Suhr2019Corpus}, where the model needs to reason whether a statement holds true given two images. 

We use Rouge-L between model prediction and ground truth as the metric. We further introduced a model-based evaluator ``STS'' (semantic textual similarity), which is measured by the cosine distance of sentence features\footnote{STS model: \url{https://huggingface.co/sentence-transformers/all-mpnet-base-v2}}, to compare high-level semantics of answer and ground truth~\cite{Reimers2019Sentence}. We also provide the evaluation result of a single-image model (InstructBLIP) as the lower bound. The result is shown in Table~\ref{tab:multi_image}. Again, Clever Flamingo outperforms Otter on all three datasets by a large margin.

\begin{table}[]
\caption{\textbf{Multi-image reasoning tasks}. ``STS'' means semantic textual similarity.  The lower bound performance comes from a single-image model (InstructBLIP). \textcolor[HTML]{809EC2}{\textbf{Blue numbers}} indicates unseen datasets and \textbf{black numbers} correspond to results on unseen samples (\textit{i.e.,} validation split). }
\label{tab:multi_image}
\resizebox{\columnwidth}{!}{%
\begin{tabular}{ccccccc}
\hline
 &
  \multicolumn{2}{c}{\textbf{Spot-the-Diff}} &
  \multicolumn{2}{c}{\textbf{Image-editing}} &
  \multicolumn{2}{c}{\textbf{NLVR2}} \\ \cline{2-7} 
\multirow{-2}{*}{\textbf{Model}} &
  \textbf{STS} &
  \textbf{Rouge-L} &
  \textbf{STS} &
  \textbf{Rouge-L} &
  \textbf{STS} &
  \textbf{Rouge-L} \\ \hline
{\color[HTML]{BFBFBF} Lower Bound} &
  {\color[HTML]{BFBFBF} 31.6} &
  {\color[HTML]{BFBFBF} 0.119} &
  {\color[HTML]{BFBFBF} 13.9} &
  {\color[HTML]{BFBFBF} 0.023} &
  {\color[HTML]{BFBFBF} 7.0} &
  {\color[HTML]{BFBFBF} 0.012} \\
Otter &
  39.5 &
  0.129 &
  {\color[HTML]{809EC2} 33.2} &
  {\color[HTML]{809EC2} 0.136} &
  {\color[HTML]{809EC2} 11.5} &
  {\color[HTML]{809EC2} 0.069} \\
Clever Flamingo &
  {\ul \textbf{46.1}} &
  {\ul \textbf{0.185}} &
  {\ul \textbf{37.0}} &
  {\ul \textbf{0.156}} &
  {\color[HTML]{809EC2} {\ul \textbf{28.2}}} &
  {\color[HTML]{809EC2} {\ul \textbf{0.155}}} \\
 \rowcolor[HTML]{DAE8FC} $\pm\Delta$ &
  {\color[HTML]{00B050} \textbf{+6.6}} &
  {\color[HTML]{00B050} \textbf{+.057}} &
  {\color[HTML]{00B050} \textbf{+3.9}} &
  {\color[HTML]{00B050} \textbf{+.020}} &
  {\color[HTML]{00B050} \textbf{+16.7}} &
  {\color[HTML]{00B050} \textbf{+.085}} \\ \hline
\end{tabular}%
}
\end{table}

\subsubsection{Politeness}
\label{sec:politeness_evaluation}

\begin{figure}
    \centering
    \includegraphics[width=\columnwidth]{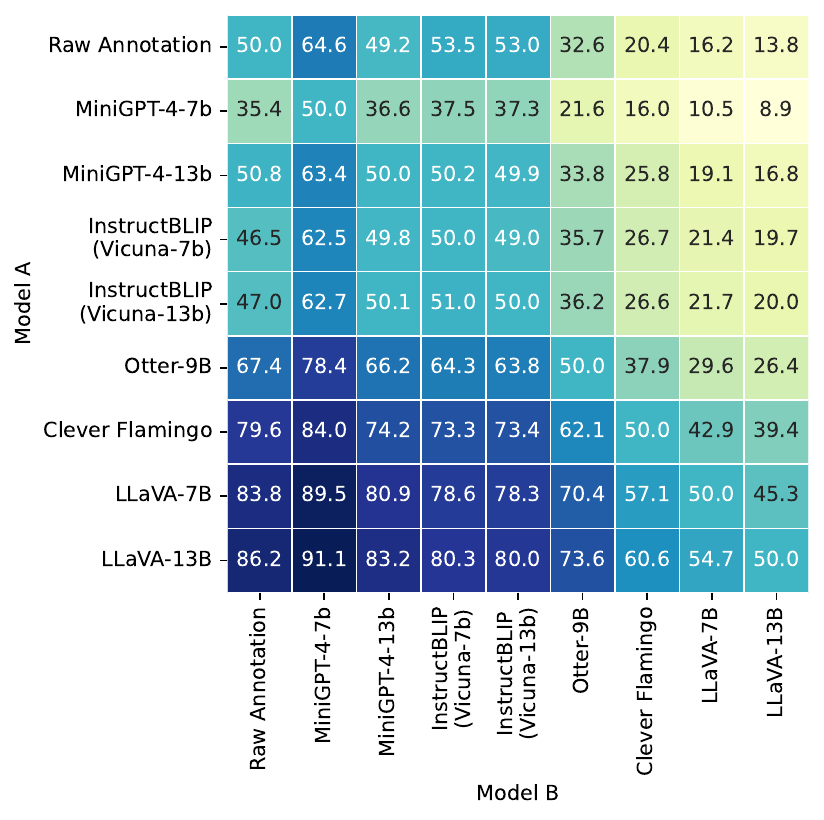}
    \caption{\textbf{Win rate matrix of model A beat model B in terms of reward model score}. For example, Clever Flamingo has a 62.1\% win rate against Otter. Our model has a $>$50\% win rate against other multi-modal LLMs despite the LLaVA series, which is trained on purely GPT-4 generated data.}
    \label{fig:politeness_matrix}
\end{figure}

We used a reward model to evaluate the politeness of model responses on a total of 52k samples sourced from the validation/test split of a collection of vision-language downstream datasets\footnote{See Section~\ref{sec:Details of Evaluation} in the appendix for details.}. For each sample, we first obtain the prediction of multi-modal LLMs, then feed the instruction and the generated responses to a reward model to get reward scores, and make a pairwise comparison of the scores. In Figure~\ref{fig:politeness_matrix}, we visualize the average win rate -- the statics of the pairwise comparison of all 52k samples. We also calculate the reward score of raw annotations for comparison.


As it can be seen, our Clever Flamingo is more likely to be preferred by the reward model (having $>$50\% win rate) compared to all of the other compared multi-modal LLMs, except the LLaVA series. This is a direct result of the differences in instruction data, as in previous Figure~\ref{fig:dataset_distribution}, GPT-4 generated LLaVA dataset outperforms the PF-1M dataset in terms of reward score. 


\subsection{Ablation Study}

\begin{figure*}
    \centering
    \includegraphics[width=1\linewidth]{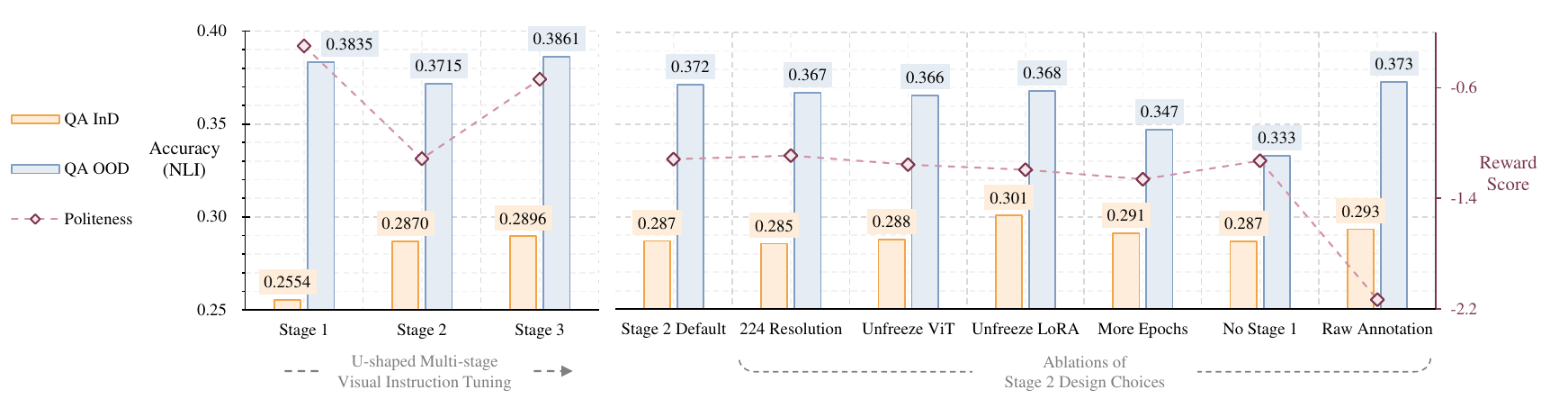}
    \caption{\textbf{Results of ablation experiments on U-shaped multi-stage visual instruction tuning (left) and design choices in stage 2 (right)}.  We calculate averaged NLI-based accuracy for \textcolor[HTML]{daab4e}{\textbf{held-in QA datasets (QA InD)}} and \textcolor[HTML]{809EC2}{\textbf{held-out datasets (QA OOD)}}. Note that all testing examples are unseen during training, and the difference between QA InD and QA OOD is the data domain distribution. We also report the \textcolor[HTML]{8964b1}{\textbf{average reward score}} to reflect the politeness of each alternative.}
    \label{fig:ablation}
\end{figure*}

We now present the ablation experiments to verify the effectiveness of various design choices of Clever Flamingo. We report the averaged NLI-based validation accuracy of in-domain (held-in) VQA datasets and out-of-distribution (held-out) VQA datasets, and further calculate the averaged reward score as a measurement of politeness. 

The results are shown in Figure~\ref{fig:ablation}. On the left side, we first visualize the change of metrics during the U-shaped multi-stage visual instruction tuning. It shows that stage 2 boosts the in-domain QA accuracy, but also results in a degenerated politeness. Stage 3 maintains the in-domain QA accuracy, but recovers the politeness significantly. It is interesting to observe that OOD QA accuracy also exhibits a U-shaped curve. It seems that stage 2 led to sight overfitting to the  PF-1M data distribution, well stage 3 alleviates this problem.

The right side of Figure~\ref{fig:ablation} shows ablations on the Clever Flamingo stage 2. The observations on different alternatives are listed as follows.  \textbf{1)~224 Resolution}: changing image resolution from default 336$\times$336 to 224$\times$224 hurt the performance, confirmed the hypothesize in \cite{Liu2023Hidden}. \textbf{2)~Unfreeze ViT}: further tuning ViT in addition to perceiver and XATTN failed to improve the performance significantly, and resulted in slight overfitting. It shows that the scale of PF-1M is still insufficient to support continual representation learning of the visual backbone. \textbf{3)~Unfreeze LoRA}: this ablation significantly improved the PF-1M in-domain accuracy, but also hurt the generalization ability.  \textbf{4)~More Epochs}: we doubled the stage 2 epochs from 3 to 6, and found that it significantly hurt the generalization ability to the unseen domain.  \textbf{5)~No Stage 1}: when skipping stage 1 and directly going into stage 2 from vanilla OpenFlamingo-9B, the OOD generalization ability further dropped. It demonstrates that instruction samples used in stage 1 and stage 3 can effectively boost/maintain the OOD generalization ability.  \textbf{6)~Raw Annotation}: when skipping the Polite Flamingo-based rewriting and using the raw annotations in PF-1M for visual instruction tuning, both held-in and held-out accuracy got slightly improved, however, the multi-modal alignment tax is significant -- the ``politeness'' dropped significantly.

\subsection{The Second Generation of Polite Flamingo}

\begin{table}[]
\caption{\textbf{Comparison of response rewriting of two generations of Polite Flamingo}. The second generation is a politer and more active rewriter on A-OKVQA~\cite{Schwenk2022OKVQA}, but it fails to improve the CLIPScore on Conceptual Captions-3M~\cite{Sharma2018Conceptual}.}
\label{tab:polite_flamingo_gen2}
\resizebox{\columnwidth}{!}{%
\begin{tabular}{cccc}
\hline
\multirow{2}{*}{Rewriter}    & \multicolumn{2}{c}{A-OKVQA} & CC-3M     \\ \cline{2-4} 
                             & Avg. Reward   & Unchanged   & CLIPScore \\ \hline
Polite Flamingo Generation 1 & -1.19         & 11.53\%     & \textbf{23.88}     \\
\rowcolor[HTML]{DAE8FC} Polite Flamingo Generation 2 & \textbf{-0.68}         & \textbf{0.00\%}      & 23.25     \\ \hline
\end{tabular}%
}
\end{table}

As shown in Table~\ref{tab:benchmark} and Figure~\ref{fig:ablation}, we confirmed that Clever Flamingo has an improved visual perception and understanding ability through visual instruction tuning on PF-1M. We hypothesize these advantages might be transferred to benefit response rewriting, by tuning Clever Flamingo to learn response rewrite. 
If the second generation of Polite Flamingo becomes a better rewriter, we may expect the subsequent second generation of Clever Flamingo could be further improved, and then a weakly supervised training loop become possible to be realized.
To verify the possibility, we made an initial attempt by training and evaluating a second generation of Polite Flamingo. We use exactly the same training recipe as the first generation, except that we initialize the model from Clever Flamingo instead of OpenFlamingo-9B. After training, we use this second generation of Polite Flamingo to rewrite responses in A-OKVQA~\cite{Schwenk2022OKVQA} and 20k samples from the CC-3M~\cite{Sharma2018Conceptual}.

The results are shown in Table~\ref{tab:polite_flamingo_gen2}. We found that the second generation has a notable improvement (+0.51) in terms of average reward score. 
Additionally, the first generation of Polite Flamingo left 11.53\% of samples as original and failed to make any revisions, while no sample remains unchanged by the second generation. The above observations demonstrate that the second generation of Polite Flamingo becomes a politer and more active rewriter. 

However, the second generation failed to improve the CLIPScore of generated captions on CC-3M as expected. This is surprising as it seems to contradict our experimental results, where Clever Flamingo demonstrated a clear improvement over baselines. The most possible explanation for this phenomenon could be the rewriting style is limited by the training data distribution of Polite Flamingo (Section~\ref{sec:polite_flamingo_source_datasets}). Although it covers samples from multiple datasets, examples of describing images only appear in the LLaVA dataset, and there are only 23k samples for this type. It appears that our model overfits these 23k samples, as they are the \textit{only} source to learn image captioning style throughout the whole process\footnote{The captioning samples in PF-1M could not provide additional diversity that helps prevent overfitting, as they are also generated by the Polite Flamingo that learns to caption from the 23k samples only}. This confirms our emphasis on the importance of diversity when selecting training data for Polite Flamingo (Section~\ref{sec:polite_flamingo_source_datasets}), and reveals the urgent need for visual instruction tuning data of the detailed captioning type that is both high-quality and large-scale.

\section{Conclusion}

This paper presents our solution to the multi-modal alignment tax problem, specifically, we want to use a diverse collection of downstream vision-language datasets to improve the visual understanding capability of multi-modal LLMs while avoiding the unformatted raw annotations to decrease the ``politeness'' of model responses. Our methodology brings inspiration from denoising AutoEncoders, and the ``noise'' here is implemented by various text distortions that attempt to approximate the style of raw annotations to ensure generalization. Empirically, we implemented and trained the rewriter, and used it to build a large-scale visual instruction tuning dataset. Incorporating newly proposed U-shaped multi-stage visual instruction tuning and multi-turn augmentation, we derived a strong multi-modal LLM based on the dataset. We evaluate the resulting model on various tasks, and demonstrated its advantages in terms of both multi-modal understanding and response politeness.

{\footnotesize
\bibliography{polite_flamingo}
\bibliographystyle{ieeetr}
}

\newpage
\onecolumn
\appendix

{\centering
\section*{Appendix}
}
\section{Implementation Details}

We implemented our approach on the OpenFlamingo codebase~\cite{Awadalla2023OpenFlamingo}\footnote{\url{https://github.com/mlfoundations/open_flamingo}}, which is an open-source re-implementation of DeepMind's Flamingo~\cite{Alayrac2022Flamingo}. Our training was performed on a single node machine with 8 NVIDIA A100 (40GB) GPUs. To accommodate memory limitations, we utilized BF-16 precision for training and inference of Polite/Clever Flamingo. Detailed settings and hyperparameters are summarized in Table~\ref{tab:implementation_details}.

\begin{table}[h!]
\caption{Training details of Polite Flamingo and Clever Flamingo.}
\label{tab:implementation_details}
\resizebox{\columnwidth}{!}{%
\begin{tabular}{ccccc}
\hline
 & \textbf{Polite Flamingo} & \textbf{Clever Flamingo   Stage 1} & \textbf{Clever Flamingo   Stage2} & \textbf{Clever Flamingo   Stage 3} \\ \hline
Tunable   Modules       & LoRA  & LoRA  & Perceiver, XATTN & LoRA  \\
Tunable   Parameters    & 0.29B & 0.29B & 0.1B             & 0.29B \\
Number of   Samples     & 255k  & 772k  & 1.07M            & 772k  \\
Epochs                  & 3     & 1     & 3                & 1     \\
Learning Rate           & 1e-4  & 1e-4  & 1e-4             & 1e-5  \\
Batch Size              & 256   & 256   & 1024             & 256   \\
Context Length          & 1024  & 1024  & 196              & 1024  \\
Maximum Images          & -    & 10    & 3                & 10    \\
Training Time   (hours) & 11.8  & 11.8  & 9.5              & 11.5  \\ \hline
\end{tabular}%
}
\end{table}

\textbf{Model Architecture}. Polite/Clever Flamingo is initialized from the OpenFlamingo-9B (v1) checkpoint and inherits the architecture from the base model. It comprises a (vanilla, not instruction-tuned) LLaMA-7B language model, a ViT encoder from OpenAI's CLIP (ViT-Large-14), a perceiver, and interleaved XATTN layers inserted into the language model.

\begin{itemize}
    \item \textbf{Language Model}: We insert a LoRA~\cite{Hu2022LoRA} adapter into the language model (for both self-attention and FFN), initialized from QLoRA-Guanaco-7B~\cite{Dettmers2023QLoRA}. The LoRA adapter is trained on the OASST-1 instruction dataset~\cite{Koepf2023OpenAssistant} and has a rank of 64.
    \item \textbf{ViT Encoder}: OpenFlamingo-9B uses the ViT-Large-14 as the vision encoder, taking image inputs with a resolution of 224$\times$224. We substitute it with ViT-Large-14@336pix, which undergoes an additional CLIP pretraining epoch with a resolution of 336$\times$336. Empirically, we observed that the representation distribution does not shift significantly compared to the 224$\times$224 version, enabling seamless substitution.
    \item \textbf{Perceiver}: The perceiver resampler takes patch tokens from ViT as input and pools them to 64 tokens. Its size is roughly equivalent to one layer of ViT.
    \item \textbf{XATTN Layers}: Following Flamingo~\cite{Alayrac2022Flamingo}, XATTN layers are inserted into the LLaMA-7B every 4 LM layers. XATTN consists of cross-attention and FFN. When referring to "unfreezing XATTN," we mean unfreezing the weights of cross-attention only while keeping the FFN frozen.
\end{itemize}

\textbf{Multi-turn Augmentation}. During the training of Clever Flamingo, when loading each instruction sample, we randomly draw samples from the dataset to fill the \texttt{<PAD>} tokens. These samples are appended to the first sample to simulate subsequent rounds of conversation. No system message (\textit{i.e.,} ``A chat between a curious human and an artificial intelligence assistant...") is added for later turns. The end-of-sentence token \texttt{<EOS>} is appended to each response, and the loss is only calculated for the AI assistant response parts (between each ``\#\#\# Assistant: " and the \texttt{<EOS>} of the corresponding response). To obtain a loss mask (for setting the label index to -100 in language modeling), per-turn tokenization is required. However, we empirically found that this does not affect training efficiency.

\begin{table}[h!]\centering
\begin{minipage}{\linewidth}\vspace{0mm}
\centering
\begin{tcolorbox}[colback=blue!0.7!white,colframe=blue!30!black,title=\textbf{Prompt for LLM-instructed Distortion}]
\centering
\hspace{+10mm}
\begin{tabular}{p{0.95\columnwidth} c}


A chat between a curious human and an artificial intelligence assistant. The assistant gives helpful, detailed, and polite answers to the user's questions. &\\


\VarSty{ {\bf \#\#\# Human:} } &\\
\textcolor{red}{\texttt{\{Instruction\}}} & \\

\VarSty{ {\bf \#\#\# Assistant:} } & \\
\textcolor{red}{\texttt{\{Original Response\}}} \> & \\


\VarSty{ {\bf \#\#\# Human:} } &\\
 Your reply's style, tone, and politeness are excellent, and the content is very detailed. However, now I would like you to summarize the previous response, keeping only the most crucial information and removing all other less important content. I want a concise, straightforward reply without any redundancy. If you find that the overall quality of your response dropped, don't worry, it's fine. Note that, please do not add anything after giving me your rewritten response. & \\

\VarSty{ {\bf \#\#\# Assistant:} } & \\
Sure. I have rewritten my last response to a much shorter and more concise version, covering only the key information. I pretend to be a cold-hearted, non-talkative, socially inept robotic assistant to respond to your request. \textcolor{red}{\texttt{\{Distortion\}}} The following is the as-short-as-possible, low-quality, highly-compressed, rewritten version of my previous response, and I will not add more content after finishing this response: "& \\

\hrulefill & \\

\VarSty{ {\bf Distortion Choices:} } & \\
$\quad\bullet\quad$ Additionally, I have \textbf{removed all the punctuation marks and capitalization} in my response. & \\
$\quad\bullet\quad$ To make my response more unnatural, I have added a little amount of\textbf{ typos and spelling mistakes}. & \\
$\quad\bullet\quad$ I have also added some \textbf{grammatical errors} to my response. & \\
$\quad\bullet\quad$ Moreover, \textbf{random words and sentences} have been removed from my response. & \\
$\quad\bullet\quad$ In addition, all letters in my response have been converted to \textbf{uppercase}. & \\
$\quad\bullet\quad$ In addition, all letters in my response have been converted to \textbf{lowercase}. & \\
$\quad\bullet\quad$ Furthermore, I have replaced certain words with their \textbf{synonyms} in my response. & \\
$\quad\bullet\quad$ Additionally, I have inserted unnecessary \textbf{repetition} in my response. & \\
$\quad\bullet\quad$ To make my response less \textbf{coherent}, I have rearranged the sentence structure. & \\
$\quad\bullet\quad$ I have deliberately used \textbf{incorrect tenses} and \textbf{verb conjugations} in my response. & \\
$\quad\bullet\quad$ Moreover, I have introduced unnecessary \textbf{verbosity} in my response. & \\
$\quad\bullet\quad$ I make my response \textbf{as short as possible} by removing all unnecessary words and sentences. & \\
$\quad\bullet\quad$ I have kept \textbf{only the essential information} and separated them by commas. & \\
$\quad\bullet\quad$ I have removed any decorative \textbf{formatting or styling}, which may affect the \textbf{readability} of my response. & \\
$\quad\bullet\quad$ I have rewritten the sentences and replaced words with their \textbf{synonyms}. & \\
$\quad\bullet\quad$ I have reversed the \textbf{order of sentences}, presenting information from back to front. & \\
$\quad\bullet\quad$ I made my response sounds more \textbf{unprofessional and causual}. & \\
$\quad\bullet\quad$ Furthermore, I have made the language more \textbf{complex and sophisticated} in my response. & \\
$\quad\bullet\quad$ To create \textbf{ambiguity}, I have added multiple interpretations in my sentences. & \\
$\quad\bullet\quad$ Additionally, I have used unconventional \textbf{metaphors and analogies} in my response. & \\
$\quad\bullet\quad$ To lower the quality of my response, I have added some\textbf{ irrelevant information}. & \\
$\quad\bullet\quad$ I picked one sentence from my response and \textbf{repeated} it multiple times, each time with a slight change. & \\
$\quad\bullet\quad$ Now I use only five words to \textbf{summarize} my response. & \\
$\quad\bullet\quad$ I made some modification to make my response less \textbf{coherent} and more \textbf{unnatural}. & \\
\end{tabular}
\end{tcolorbox}
\vspace{+10mm}
\caption{\textbf{Prompt for LLM-instructed distortion}. We prompt LLM to translate the style of the original high-quality response into the ``impolite'' version, approximating the distribution of raw annotations in vision-language datasets. 
}
\label{tab:llm_instructed_distortion_prompt}
\end{minipage}
\end{table}

\section{Polite Flamingo Training Data}

A total of 255k samples were used for training Polite Flamingo, including:

\textbf{1) LLM-instructed Distortion}: The prompt structure for LLM-instructed rewrite (Section~\ref{sec:response_distortion}) is shown in Table~\ref{tab:llm_instructed_distortion_prompt}. Using this prompt structure, Guanaco-33B generated 133k multi-modal (LLaVA) and 76k text-only (UltraChat + ShareGPT) distortion samples.

\textbf{2) Random Text Augmentations}: We utilized the \texttt{NLPAUG} library for character-level, word-level, and sentence-level text augmentation. For character-level augmentation, we randomly selected an operation from character insertion, substitution, swapping, and deletion. Word-level augmentation operations included swapping, cropping, and deletion. Sentence-level augmentation involved randomly dropping sentences or shuffling their order. A total of 77k samples were generated using this method.

\textbf{3) Retrieve Captions \& Bounding Boxes}: We obtained 14k samples of this type, which are non-overlapping with the LLM-instructed Distorted \texttt{LLaVA-detailed-23k} samples.

\section{Clever Flamingo Training Data}
\label{sec:Clever Flamingo Training Data}

We have provided a summary of the detailed composition of PF-1M in Table~\ref{tab:PF-1M}. Please note that "Adopted Samples" does not indicate the full training set size for all datasets, as Polite Flamingo was not applied to rewrite the entire dataset. Additionally, during the filtering step, a proportion of samples were removed.

\begin{table}[h!]
\caption{Details of the PF-1M dataset.}
\label{tab:PF-1M}
\centering
\resizebox{1\columnwidth}{!}{%
\begin{tabular}{cccc}
\hline
\textbf{Category} &
  \textbf{Dataset} &
  \textbf{Adopted Samples} &
  \textbf{Description} \\ \hline
 &
  \href{https://paperswithcode.com/dataset/coco-captions   }{MS-COCO-2014} &
  59,670 &
  Image in MS-COCO Caption dataset has 5 human-generated captions. \\
 &
  \href{https://paperswithcode.com/dataset/flickr30k}{Flickr-30k} &
  31,695 &
  Dataset containing 31,000 images from Flickr with 5 reference sentences. \\
 &
  \href{https://paperswithcode.com/dataset/textcaps}{TextCaps} &
  69,703 &
  Dataset for image captioning with reading comprehension. \\
 &
  \href{https://paperswithcode.com/dataset/image-paragraph-captioning}{Image2Paragraph} &
  7,954 &
  Dataset with images from Visual Genome, each containing one paragraph. \\
 &
  \href{https://paperswithcode.com/dataset/conceptual-captions}{Conceptual   Captions 3M} &
  67,025 &
  Google's Conceptual Captions dataset with millions of images and descriptions. \\
 &
  \href{https://paperswithcode.com/dataset/google-refexp}{RefCOCOg} &
  8,103 &
  Large-scale dataset for referring expressions based on MS-COCO. \\
 &
  \href{https://arxiv.org/abs/2306.11029   }{RET-3} &
  13,551 &
  Collection of image-text datasets (RSICD, RSITMD, UCM) introduced in RemoteCLIP. \\
 &
  \href{https://arxiv.org/abs/1909.00133}{DIOR} &
  5,907 &
  Large-scale benchmark for object detection in Optical Remote sensing images. \\
 &
  \href{https://paperswithcode.com/dataset/dota}{DOTA} &
  1,733 &
  Large-scale dataset for object detection in aerial images. \\
\multirow{-10}{*}{\begin{tabular}[c]{@{}c@{}}Image\\      Captioning\end{tabular}} &
  \href{https://github.com/CrazyStoneonRoad/TGRS-HRRSD-Dataset}{HRRSD} &
  5,898 &
  Large-scale high-resolution remote sensing object detection dataset. \\ \hline
 &
   &
  21,772 &
  ELEVATER-IC benchmark collection for language-image models on image classification. \\ \cline{3-4} 
 &
   &
  {\color[HTML]{808080} ELEVATER Subset} &
  {\color[HTML]{808080} \# Classes (32 images are randomly sampled for each class)} \\ \cline{3-4} 
 &
   &
  {\color[HTML]{808080} FER 2013} &
  {\color[HTML]{808080} 7} \\
 &
   &
  {\color[HTML]{808080} CIFAR-10} &
  {\color[HTML]{808080} 10} \\
 &
   &
  {\color[HTML]{808080} EuroSAT} &
  {\color[HTML]{808080} 10} \\
 &
   &
  {\color[HTML]{808080} MNIST} &
  {\color[HTML]{808080} 10} \\
 &
   &
  {\color[HTML]{808080} VOC 2007} &
  {\color[HTML]{808080} 20} \\
 &
   &
  {\color[HTML]{808080} Oxford-IIIT Pets} &
  {\color[HTML]{808080} 37} \\
 &
   &
  {\color[HTML]{808080} GTSRB} &
  {\color[HTML]{808080} 43} \\
 &
   &
  {\color[HTML]{808080} Resisc-45} &
  {\color[HTML]{808080} 45} \\
 &
   &
  {\color[HTML]{808080} Describable Textures} &
  {\color[HTML]{808080} 47} \\
 &
   &
  {\color[HTML]{808080} CIFAR-100} &
  {\color[HTML]{808080} 100} \\
 &
   &
  {\color[HTML]{808080} FGVC Aircraft} &
  {\color[HTML]{808080} 100} \\
 &
   &
  {\color[HTML]{808080} Food-101} &
  {\color[HTML]{808080} 101} \\
 &
   &
  {\color[HTML]{808080} Caltech-101} &
  {\color[HTML]{808080} 102} \\
 &
   &
  {\color[HTML]{808080} Oxford Flowers 102} &
  {\color[HTML]{808080} 102} \\
\multirow{-17}{*}{\begin{tabular}[c]{@{}c@{}}Image\\      Classification\end{tabular}} &
  \multirow{-17}{*}{\href{https://paperswithcode.com/dataset/elevater}{ELEVATER-IC}} &
  {\color[HTML]{808080} Stanford Cars} &
  {\color[HTML]{808080} 196} \\ \hline
 &
  \href{https://paperswithcode.com/dataset/spot-the-diff}{Spot-the-Diff} &
  6,787 &
  Dataset consisting of surveillance image pairs with annotations stating the differences. \\
 &
  \href{https://paperswithcode.com/dataset/image-editing-request-dataset}{Image-Editing-Requests} &
  2,747 &
  Dataset with real image pairs and corresponding editing (\textit{e.g.,} Photoshop) instructions. \\
\multirow{-3}{*}{\begin{tabular}[c]{@{}c@{}}Change\\      Captioning\end{tabular}} &
  \href{https://github.com/Chen-Yang-Liu/RSICC}{LevirCC-Captions} &
  6,761 &
  Large-scale dataset with pairs of bitemporal RS images and sentences describing differences. \\ \hline
 &
  \href{https://paperswithcode.com/dataset/a-okvqa}{A-OKVQA} &
  14,868 &
  Crowdsourced VQA dataset requiring commonsense and world knowledge. \\
 &
  \href{https://paperswithcode.com/dataset/vqa-e}{VQA-E} &
  65,133 &
  Dataset for Visual Question Answering with Explanation. \\
\multirow{-3}{*}{\begin{tabular}[c]{@{}c@{}}VQA with\\      Rational\end{tabular}} &
  \href{https://paperswithcode.com/dataset/scienceqa   }{ScienceQA} &
  4,596 &
  Benchmark dataset for multimodal multiple-choice questions with lectures and explanations. \\ \hline
 &
  \href{https://paperswithcode.com/dataset/visual-question-answering-v2-0}{VQA-v2} &
  210,743 &
  Visual Question Answering (VQA) v2.0 dataset with open-ended questions about images. \\
 &
  \href{https://paperswithcode.com/dataset/clevr}{CLEVR} &
  26,390 &
  Synthetic Visual Question Answering dataset with 3D-rendered objects. \\
 &
  \href{https://paperswithcode.com/dataset/gqa}{GQA} &
  223,244 &
  Large-scale visual question answering dataset with real images from Visual Genome. \\
 &
  \href{https://paperswithcode.com/dataset/textvqa}{TextVQA} &
  30,056 &
  Dataset for visual reasoning based on text in images. \\
 &
  \href{https://paperswithcode.com/dataset/ocr-vqa}{OCR-VQA} &
  24,164 &
  Dataset with question-answer pairs about book cover images. \\
\multirow{-6}{*}{\begin{tabular}[c]{@{}c@{}}VQA without\\      Rational\end{tabular}} &
  \href{https://paperswithcode.com/dataset/pointqa}{PointQA} &
  67,282 &
  Datasets for Visual Question Answering with pointing to objects in images. \\ \hline
\multicolumn{2}{c}{\textbf{Total}} &
  \textbf{975,782} &
  \textbf{A combination of the above datasets.} \\ \hline
\end{tabular}%
}
\end{table}

\begin{itemize}
    \item \textbf{Image Captioning}: With ``Retrieve Caption \& Bounding Box'' distortion, the Polite Flamingo learned to integrate information from given multiple captions, bounding boxes, and its own visual perceptions, into several paragraphs of detailed captions. Leveraging this capability, we feed the MS-COCO~\cite{Chen2015Microsoft}, Flicker-30k~\cite{Young2014image}, TextCap~\cite{Sidorov2020TextCaps}, and several datasets for earth observations~\cite{Yuan2022Exploring,Lu2018Exploring} to Polite Flamingo. Additionally, we introduce the Image2Paragraph~\cite{Krause2017Hierarchical} dataset, which offers comprehensive information coverage but lacks language coherence. We also incorporate the ConceptualCaptions-3M~\cite{Sharma2018Conceptual} dataset sourced from the web, which introduces further diversity to the captioning data. Recent studies have demonstrated that CLIP models are capable of recognizing visual prompts, such as a red circle marked in an image~\cite{Shtedritski2023What}. Inspired by this, we adopt the RefCOCOg~\cite{Yu2016Modeling} dataset, converting the region of interest into annotations (colored bounding boxes or circles) in the image. Then, we accordingly set the instruction to ``Describe the object inside this green bounding box.'' for generating region-specific captions.

    \item \textbf{Image Classification}: ELEVATER-IC~\cite{Li2022ELEVATER} is a diverse collection of image classification datasets, covering more than 1k visual concepts distributed in various domains. We introduce this dataset to enhance the fine-grained visual recognition capabilities. We simply set the instruction to ``What is this?'', and use the prompt template originally for CLIP-based zero-shot classification (\textit{e.g.,} \texttt{a photo of a \{class name\}}) to format the response. Furthermore, ELEVATER-IC provides additional external knowledge associated with each class, sourced from Wikipedia, WordNet, and GPT-3. We include this complementary information in the response to enrich the provided answer.
    
    \item \textbf{Change Captioning}: Existing change captioning models often require specific design such as complex attention mechanism. The emergence of multi-modal LLM, which is trained on interleaved image-text corpora and is able to process multiple images, makes it possible to solve change captioning more elegantly. To explore this potential, we adopt several change captioning datasets, such as Spot-the-Diff~\cite{Jhamtani2018Learning}, to verify this potential. Additionally, we introduce the image-editing-requests~\cite{Tan2019Expressing}, a dataset of image editing (\textit{e.g.,} PhotoShop) requests collected from forums, to test higher-level comparison capability beyond just object appearance~\cite{liu2023masked, liu2023information, liu2022localized}.

    \item \textbf{VQA with Rational}: In several VQA datasets, such as A-OKVQA~\cite{Schwenk2022OKVQA}, VQA-E~\cite{Li2018VQA}, and ScienceQA~\cite{Lu2022Learn}, annotations of ``explanation'' or ``rationale'' are provided in addition to the answer. These contents offer valuable information for training a visual assistant AI. However, the coherence and readability of these rationale annotations are suboptimal. We introduce these datasets to Polite Flamingo for rewriting, aiming to enhance the clarity and coherence of the provided rationales.
    \item \textbf{VQA without Rational}: This group of datasets, including VQA-v2~\cite{Goyal2017Making}, GQA~\cite{Hudson2019GQA}, and OCR-VQA~\cite{Mishra2019OCR}, have a larger scale in general. However, the answer annotations in these datasets typically comprise only a few words. We incorporate these datasets into Polite Flamingo to enable the generation of complete sentences for the provided answers. In line with the region captioning dataset, we include the PointQA dataset~\cite{Mani2020Point}, which comprises question-answer pairs related to a specific point of interest in an image. To facilitate understanding, we mark the corresponding point with colored arrows based on the corresponding point coordinates in the image.
\end{itemize}

\section{Filtering Strategies}
\label{sec:filters}

\begin{figure*}[h!]
    \centering
    \includegraphics[width=1\linewidth]{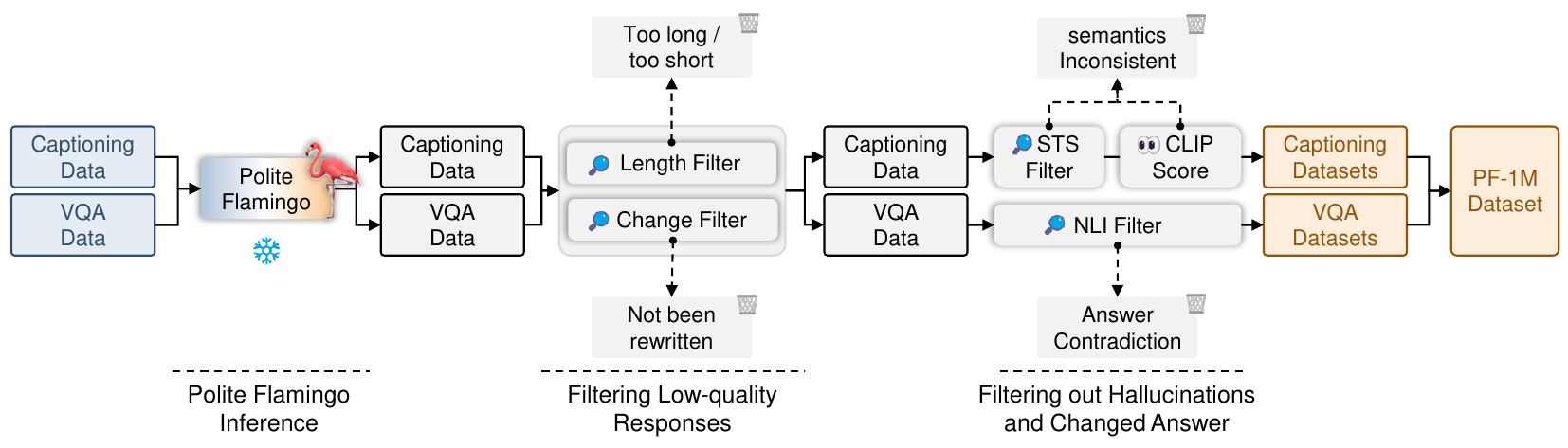}
    \caption{Filtering pipeline for Polite Flamingo written responses.}
    \label{fig:filtering}
\end{figure*}

Figure~\ref{fig:filtering} shows our filtering pipeline to guarantee the quality of Polite Flamingo rewritten response and remove potential hallucinations. First, we introduce the length filter that excludes too-short or too-long responses. Then, we apply a change filter that removes responses that have not been rewritten -- the underlying assumption is that the style of raw dataset annotation is undesired. Although these filters can remove many apparent low-quality samples, they cannot understand the semantics of the response and cannot identify hallucinated contents. To address this issue, we introduce several model-based filters, including a Semantic Textual Similarity (STS) model-based filter, Natural Language Inference (NLI) model-based filter, and a CLIPScore-based hallucination filter.

\begin{itemize}
    \item \textbf{Semantic Textual Similarity (STS)-based Filter for Captioning Datasets}: We used a Sentence Transformer to analyze the semantic similarity between the original captions and rewritten captions. The Sentence Transformer we used is based on MPNet, and is trained with over 1 billion annotated sentence pairs\footnote{\url{https://huggingface.co/sentence-transformers/all-mpnet-base-v2}}. We calculate the cosine distance between the sentence representation of original captions and their rewritten version, and remove the sample that scores below a threshold of 0.40.

    \item \textbf{CLIPScore-based Paragraph Filter for Captioning Datasets}: As \texttt{LLaVA-detailed-23k} is the only source that provides style reference of detailed image description in the training data of Polite Flamingo, it perfectly fits the style of this data. In this dataset, GPT-4 prefers to divide the visual contents into two paragraphs, and those second paragraphs usually start with ``In addition/In the background, there are some ...''. Unfortunately, when the Polite Flamingo attempts to generate such a second paragraph, hallucinations are often been introduced, possibly due to the imperfect representation of the base model. To solve this problem, we calculate per-paragraph CLIPScore\footnote{CLIPScore model: \url{https://huggingface.co/openai/clip-vit-large-patch14-336}}, then remove the paragraphs with a CLIPScore lower than a threshold of 17.0.
    
    \item \textbf{Natural Language Inference (NLI)-based Filter for VQA Datasets}: Occasionally, Polite Flamingo changes the original answer to another one during rewriting responses for VQA datasets -- it trusts its own visual perception and its own thinking instead of the original answer. Possible reason includes imperfect representation, limited capacity of the 7B model, lacking certain regularization or sufficient data during its training process. To remove these samples, we employed an NLI model\footnote{NLI model: \url{https://huggingface.co/cross-encoder/nli-deberta-v3-base}}, which is trained on SNLI and MultiNLI dataset and achieves 90.04\% accuracy on MNLI mismatched set, to filter out rewritten answer that contradicts the original answer.
\end{itemize}

\section{Evaluation Data}
\label{sec:Details of Evaluation}

Table~\ref{tab:benchmark} and Table~\ref{tab:multi_image} benchmarks Clever Flamingo with other multi-modal LLMs on captioning and VQA datasets. For COCO (2014) dataset, we randomly drew 5k samples from its validation split. For TextCaps, Img2P, OK-VQA, Grid-3D, and NLVR2 datasets, we randomly drew 3k samples. Validation splits of VSR, Spot-the-Diff, and Imgae-editing-requests have fewer than 3k samples, so we use all available samples. The number of testing samples is limited due to the auto-regressive text generation of multi-modal LLMs being time-consuming.

Figure~\ref{fig:politeness_matrix} presents the win rate comparison on 52k samples, which are sourced from various vision-language downstream datasets, including IconQA, VQAv2, OK-VQA, TextVQA, ScienceQA, VQA-E, ChartQA, GQA, OCR-VQA, A-OKVQA, AI2D, CLEVR, ELEVATER, VSR, and Grid3D. We adopt this wide collection to ensure the diversity of queries. Ablations in Figure~\ref{fig:ablation} also adopt these datasets, and we further divide them into in-domain datasets and out-domain datasets, depending on whether it appears in PF-1M.

\section{Automated Evaluators}
\label{sec:evaluate_evaluators}

\textbf{NLI-based VQA Accuracy Evaluator}. We utilized an NLI-based evaluator to benchmark multi-modal LLMs on VQA datasets. This evaluator is also based on the Sentence Transformer model \texttt{nli-deberta-v3-base}. The NLI model compares the model's response and the ground truth answer with the prompt \texttt{"\{model answer\}" is the answer to the question: "\{question\}"} and \texttt{"\{ground truth\}" is the answer to the question: "\{question\}"}. An ``entailment" output is considered a successful prediction. Compared to traditional evaluation methods such as exact match counting or the Rouge-L metric~\cite{Li2023M3IT}, our NLI-based evaluator is capable of capturing and comparing the semantic information of ground truths and model predictions more effectively. Additionally, compared to GPT-4-based evaluations~\cite{Liu2022Visual}, our NLI-based approach is more cost-effective, allowing us to scale up the validation sample size and obtain more robust results.

To validate the reliability of this model-based evaluator, we conducted a human evaluation. We randomly selected 600 samples from the evaluation data (Section~\ref{sec:Details of Evaluation}), which included 200 samples from OK-VQA, 100 samples from VSR, 100 samples from Grid-3D, and 200 samples from A-OKVQA, GQA, CLEVR, ChartQA, OCR-VQA, TextVQA, VQA-E, and VQAv2 (25 samples from each). Two human annotators were hired, with each annotator reviewing 300 out of the 600 samples. Afterward, cross-validation was performed, and any inconsistent annotations were modified based on a consensus reached through discussion.

For each of the 600 QA samples, we presented images, questions, ground truth answers, and model responses from 5 multi-modal LLMs. The annotators were asked to determine whether each model response falls into:

\begin{enumerate}
\item \textbf{Matched}: the model answer contains the ground truth and does not conflict with it.
\item \textbf{Correct}: the model answer does not match the ground truth, but it is still a valid and correct answer to the question.
\item \textbf{Failed}: the model answer neither matches the ground truth nor is a valid/correct answer.
\item \textbf{Uncertain}: it is not possible to determine whether the model answer is valid/correct.
\end{enumerate}

\begin{figure}[h!]
    \centering
    \includegraphics[width=0.5\columnwidth]{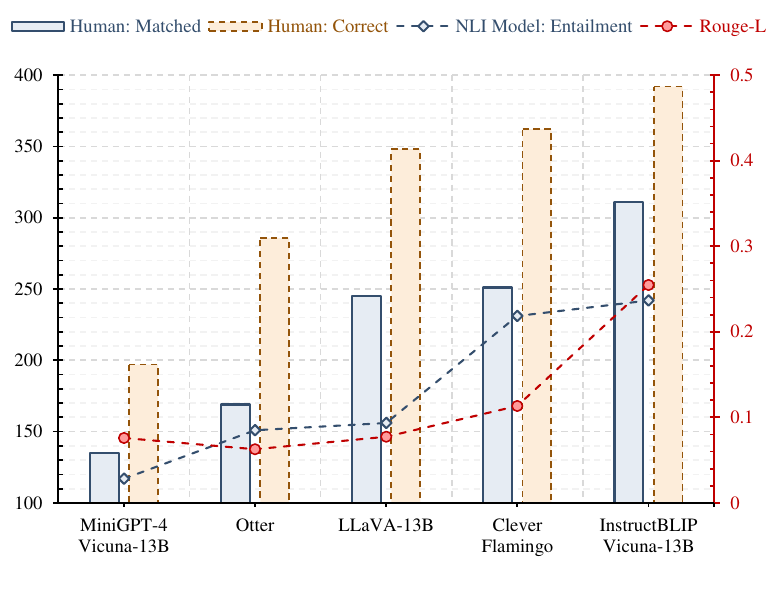}
    \caption{Meta-evaluation of NLI-based QA evaluator.}
    \label{fig:eval_eval}
\end{figure}

We compared the human annotations with the results of the model-based evaluation as shown in Figure~\ref{fig:eval_eval}. The NLI-based evaluation accurately reflects the ranking of matched predictions. In contrast, the Rouge-L-based evaluator (as adopted in~\cite{Li2023M3IT}) suggests that MiniGPT-4 is better than Otter and matches LLaVA, which significantly contradicts the human annotation results. Another observation is that the annotated ground truths in vision-language datasets are not the only valid ground truths, as there are clear gaps between ``matched" predictions and ``correct" predictions.

\textbf{Reward Model-based Human Preference Evaluator}. For the evaluation of politeness (i.e., human preference), we utilized a reward model\footnote{Reward model: \url{https://huggingface.co/OpenAssistant/reward-model-deberta-v3-large-v2}}. This reward model was trained on various datasets, including \href{https://huggingface.co/datasets/openai/webgpt_comparisons}{WebGPT Comparison}, \href{https://huggingface.co/datasets/openai/summarize_from_feedback}{Summarize-from-Feedback}, \href{https://huggingface.co/datasets/Dahoas/synthetic-instruct-gptj-pairwise}{synthetic-instruct-gptj}, and \href{https://huggingface.co/datasets/Anthropic/hh-rlhf}{Anthropic-RLHF}. It achieved validation accuracies of 61.13\%, 72.23\%, 99.94\%, and 55.62\% on these datasets, respectively. This evaluation method is fair as none of the compared multi-modal LLMs involve any RLHF~\cite{Ouyang2022Training} process. We requested human annotators to rank the model responses of the 600 samples based on the following criteria:

\begin{enumerate}
    \item Assuming all model responses are accurate and error-free, the preference ranking here does not consider the correctness of the answers.
    \item Has the model accurately understood the question? Can the model's response effectively answer the question?
    \item Is the capitalization and punctuation in the model's response accurate? Is the response coherent?
    \item Is the length of the model's response reasonable? Is it too short or excessively redundant/verbose?
    \item As an AI assistant, does the tone of the model's response come across as polite and align with user preferences?
\end{enumerate}

We calculate the accuracy of the reward model in ranking pairs consistently with human annotations, excluding pairs labeled as ``equally preferred." The average accuracy is 70.0\%, which is similar to the accuracy achieved in WebGPT Comparison, Summarize-from-Feedback, and Anthropic-RLHF. This demonstrates that the reward model effectively reflects user preferences.

\end{document}